\documentclass[wcp]{jmlr}
\usepackage{longtable}
\usepackage{booktabs}
\usepackage[nolist]{acronym}
\usepackage{algorithm,algorithmic}
\usepackage{times}
\usepackage{enumerate}

\newtheorem{assumption}{Assumption}
\newtheorem{property}{Property}

\jmlrvolume{95}
\jmlryear{2018}

\title[ASVRG: Accelerated Proximal SVRG]{ASVRG: Accelerated Proximal SVRG}

  \author{\Name{Fanhua Shang} \Email{fhshang@xidian.edu.cn}\\
  \Name{Licheng Jiao} \Email{lchjiao@mail.xidian.edu.cn}\\
  \addr Key Laboratory of Intelligent Perception and Image Understanding of Ministry of Education, School of Artificial Intelligence, Xidian University, China
  \AND
  \Name{Kaiwen Zhou} \Email{kwzhou@cse.cuhk.edu.hk}\\
  \Name{James Cheng} \Email{jcheng@cse.cuhk.edu.hk}\\
  \addr Department of Computer Science and Engineering, The Chinese University of Hong Kong, Hong Kong
  \AND
  \Name{Yan Ren} \Email{crane\_rock@outlook.com}\\
  \Name{Yufei Jin} \Email{jesty@jestyf.cn}\\
  \addr School of Computer Science and Technology, Xidian University, China
 }

\editors{Jun Zhu and Ichiro Takeuchi}

\begin{document}

\maketitle

\begin{abstract}
This paper proposes an accelerated proximal stochastic variance reduced gradient (ASVRG) method, in which we design a simple and effective momentum acceleration trick. Unlike most existing accelerated stochastic variance reduction methods such as Katyusha, ASVRG has only one additional variable and one momentum parameter. Thus, ASVRG is much simpler than those methods, and has much lower per-iteration complexity. We prove that ASVRG achieves the best known oracle complexities for both strongly convex and non-strongly convex objectives. In addition, we extend ASVRG to mini-batch and non-smooth settings. We also empirically verify our theoretical results and show that the performance of ASVRG is comparable with, and sometimes even better than that of the state-of-the-art stochastic methods.
\end{abstract}


\section{Introduction}
Consider the following composite convex minimization:
\begin{equation}\label{equ1}
\min_{x\in\mathbb{R}^{d}} F(x):=f(x)+g(x)=\frac{1}{n}\!\sum_{i=1}^{n}\!f_{i}(x)+g(x),
\end{equation}
where $f(x)\!:=\!\frac{1}{n}{\textstyle\sum}^{n}_{i=1}f_{i}(x)$ is a convex function that is a finite average of $n$ convex component functions $f_{i}(x)\!:\!\mathbb{R}^{d}\!\rightarrow\!\mathbb{R}$, and $g(x)$ is a ``simple" possibly non-smooth convex function. This formulation naturally arises in many problems in machine learning, optimization and signal processing, such as regularized empirical risk minimization (ERM)) and eigenvector computation \citep{shamir:pca,garber:svd}. To solve Problem (\ref{equ1}) with a large sum of $n$ component functions, computing the full gradient of $f(x)$ in first-order methods is expensive, and hence stochastic gradient descent (SGD) has been widely applied to many large-scale problems \citep{zhang:sgd,krizhevsky:deep}. The update rule of proximal SGD is
\begin{equation}\label{equ2}
x_{t}=\mathop{\arg\min}_{y\in\mathbb{R}^{d}}\, \left\{\frac{1}{2\eta_{t}}\|y-x_{t-1}\|^2+y^{T}\nabla\! f_{i_{t}}\!(x_{t-1})+g(y)\right\},
\end{equation}
where $\eta_{t}\!\propto\!1/\sqrt{t}$ is the step size, and $i_{t}$ is chosen uniformly at random from $\{1,\ldots,n\}$. When $g(x)\!\equiv\!0$, the update rule in~\eqref{equ2} becomes $x_{t}\!=\!x_{t-1}\!-\!\eta_{t}\nabla\! f_{i_{t}}\!(x_{t-1})$. The standard SGD estimates the gradient from just one example (or a mini-batch), and thus it enjoys a low per-iteration cost as opposed to full gradient methods. The expectation of $\nabla\!f_{i}(x_{t})$ is an unbiased estimation to $\nabla\!f(x_{t})$, i.e., $\mathbb{E}[\nabla\!f_{i}(x_{t})]\!\!=\!\!\nabla\! f(x_{t})$. However, the variance of the stochastic gradient estimator may be large, which leads to slow convergence \citep{johnson:svrg}. Even under the strongly convex (SC) and smooth conditions, standard SGD can only attain a sub-linear rate of convergence \citep{rakhlin:sgd, shamir:sgd}.

Recently, the convergence rate of SGD has been improved by many variance reduced SGD methods \citep{roux:sag, shalev-Shwartz:sdca, johnson:svrg, defazio:saga, mairal:miso} and their proximal variants \citep{schmidt:sag, xiao:prox-svrg, shalev-Shwartz:acc-sdca}. The methods use past gradients to progressively reduce the variance of stochastic gradient estimators, so that a constant step size can be used. In particular, these variance reduced SGD methods converge linearly for SC and Lipschitz-smooth problems. SVRG \citep{johnson:svrg} and its proximal variant, Prox-SVRG \citep{xiao:prox-svrg}, are particularly attractive because of their low storage requirement compared with the methods in \citep{roux:sag, defazio:saga, shalev-Shwartz:acc-sdca}, which need to store all the gradients of the $n$ component functions $f_{i}(\cdot)$ (or dual variables), so that $O(nd)$ storage is required in general problems. At the beginning of each epoch of SVRG and Prox-SVRG, the full gradient $\nabla\! f(\widetilde{x})$ is computed at the past estimate $\widetilde{x}$. Then the key update rule is given by
\begin{equation}\label{equ3}
\begin{split}
&\quad\,\widetilde{\nabla}\! f_{i_{t}}\!(x_{t-1})=\nabla\! f_{i_{t}}\!(x_{t-1})-\nabla\! f_{i_{t}}\!(\widetilde{x})+\nabla\! f(\widetilde{x}),\\
&x_{t}=\mathop{\arg\min}_{y\in\mathbb{R}^{d}}\, \left\{\frac{1}{2\eta}\|y-x_{t-1}\|^2+y^{T}\widetilde{\nabla}\! f_{i_{t}}\!(x_{t-1})+g(y)\right\}.
\end{split}
\end{equation}

For SC problems, the oracle complexity (total number of component gradient evaluations to find $\varepsilon$-suboptimal solutions) of most variance reduced SGD methods is $\mathcal{O}\!\left((n\!+\!{L}/{\mu})\log({1}/{\varepsilon})\right)$, when each $f_{i}(x)$ is $L$-smooth and $g(x)$ is $\mu$-strongly convex. Thus, there still exists a gap between the oracle complexity and the upper bound in \citep{woodworth:bound}. In theory, they also converge slower than accelerated deterministic algorithms (e.g., FISTA \citep{beck:fista}) for non-strongly convex (non-SC) problems, namely $\mathcal{O}(1/t)$ vs.\ $\mathcal{O}(1/t^2)$.

Very recently, several advanced techniques were proposed to further speed up the variance reduction SGD methods mentioned above. These techniques mainly include the Nesterov's acceleration technique \citep{nitanda:svrg,lin:vrsg,murata:dual,lan:rpdg}, the projection-free property of the conditional gradient method \citep{hazan:svrf}, reducing the number of gradient calculations in the early iterations \citep{babanezhad:vrsg,zhu:univr,shang:fsvrg}, and the momentum acceleration trick \citep{hien:asmd,zhu:Katyusha,zhou:fsvrg,shang:vrsgd1}. \cite{lin:vrsg} and \cite{frostig:sgd} proposed two accelerated algorithms with improved oracle complexity of $\mathcal{O}((n+\sqrt{n{L}/{\mu}})\log({L}/{\mu})\log({1}/{\varepsilon}))$ for SC problems. In particular, Katyusha \citep{zhu:Katyusha} attains the optimal oracle complexities of $\mathcal{O}((n\!+\!\!\sqrt{n{L}/{\mu}})\log({1}/{\varepsilon}))$ and $\mathcal{O}(n\log({1}/{\varepsilon})\!+\!\sqrt{nL/\varepsilon})$ for SC and non-SC problems, respectively. The main update rules of Katyusha are formulated as follows:
\begin{equation}\label{equ4}
\begin{split}
&x_{t}=y_{t-1}+\omega_{1}(z_{t-1}-y_{t-1})+\omega_{2}(\widetilde{x}-y_{t-1}),\\
&y_{t}=\mathop{\arg\min}_{y\in\mathbb{R}^{d}}\, \left\{\frac{3L}{2}\|y-x_{t}\|^2+y^{T}\widetilde{\nabla}\! f_{i_{t}}\!(x_{t})+g(y)\right\},\\
&z_{t}=\mathop{\arg\min}_{z\in\mathbb{R}^{d}}\, \left\{\frac{1}{2\eta}\|z-z_{t-1}\|^2+z^{T}\widetilde{\nabla}\! f_{i_{t}}\!(x_{t})+g(z)\right\},
\end{split}
\end{equation}
where $\omega_{1},\omega_{2}\!\in\![0,1]$ are two parameters for the key momentum terms, and $\eta\!=\!1/(3\omega_{1}L)$. Note that the parameter $\omega_{2}$ is fixed to $0.5$ in \citep{zhu:Katyusha} to avoid parameter tuning.

\textbf{Our Contributions.} In spite of the success of momentum acceleration tricks, most of existing accelerated methods including Katyusha require at least two auxiliary variables and two corresponding momentum parameters (e.g., for the Nesterov's momentum and Katyusha momentum in~\eqref{equ4}), which lead to complicated algorithm design and high per-iteration complexity. We address the weaknesses of the existing methods by a simper accelerated proximal stochastic variance reduced gradient (ASVRG) method, which requires only one auxiliary variable and one momentum parameter. Thus, ASVRG leads to much simpler algorithm design and is more efficient than the accelerated methods. Impressively, ASVRG attains the same low oracle complexities as Katyusha for both SC and non-SC objectives. We summarize our main contributions as follows.

\begin{itemize}
\item {We design a simple momentum acceleration trick to accelerate the original SVRG. Different from most accelerated algorithms such as Katyusha, which require two momentums mentioned above, our update rule has only one momentum accelerated term.}
\item {We prove that ASVRG converges to an $\varepsilon$-minimizer with the oracle complexity of $\mathcal{O}((n\!+\!\sqrt{n{L}/{\mu}})\log({1}/{\varepsilon}))$ for SC problems, which is the same as that in \citep{defazio:sagab,zhu:Katyusha}, and matches the upper bound in \citep{woodworth:bound}.}
\item {We also prove that ASVRG achieves the optimal convergence rate of $\mathcal{O}(1/t^2)$ and the oracle complexity of $\mathcal{O}(n\log({1}/{\varepsilon})\!+\!\sqrt{nL/\varepsilon})$ for non-SC problems, which is identical to the best known result in \citep{hien:asmd,zhu:Katyusha}.}
\item {Finally, we introduce mini-batching, adaptive regularization and smooth techniques into our algorithms, and further analyze their convergence properties and summarize the oracle complexities of ASVRG for the four cases of Problem~\eqref{equ1} in Table~\ref{tab1}.}
\end{itemize}

\begin{table*}[h]
\centering
\small
\caption{Comparison of the oracle complexities of some stochastic variance reduced algorithms for the four classes of problems in Section~\ref{sect20}. Note that $C_{1}\!\!=\!\|\widetilde{x}^{0}\!-\!x^{\star}\|$ and $C_{2}\!\!=\!F(\widetilde{x}^{0})\!-\!F(x^{\star})$.}
\label{tab1}
\renewcommand\arraystretch{1.65}
\begin{tabular}{l|cc|cc}
\hline
  & \multicolumn{2}{|c|}{$L$-Smooth} & \multicolumn{2}{c}{$G$-Lipschitz}\\
\hline
   & \!\!\!\!$\mu$-Strongly Convex & \!\!\!\!\!\!\!non-SC    & $\mu$-Strongly Convex  & \!\!\!\!\!\!\!non-SC  \!\!\!\! \\
\hline
\!\!SVRG\!\! & \!\!\!\!$\mathcal{O}\!\left((n\!+\!{L}/{\mu})\log\!\frac{C_{2}}{\varepsilon}\right)$ & \!\!\!\!\!\!\! unknown & \!\!\!\!  unknown & \!\!\!\!\!\!\! unknown\!\!\!\! \\
\!\!SAGA\!\!  & \!\!\!\!$\mathcal{O}\!\left((n\!+\!{L}/{\mu})\log\!\frac{C_{2}}{\varepsilon}\right)$  & \!\!\!\!\!\!\! $\mathcal{O}\!\left({nC_{2}}/{\varepsilon}\!+\!{LC^{2}_{1}}/{\varepsilon}\right)$ & \!\!\!\! unknown & \!\!\!\!\!\!\! unknown\!\!\!\! \\
\!\!Katyusha\!\!  & \!\!$\mathcal{O}\!\left((n\!+\!\!\sqrt{n{L}/{\mu}})\log\!\frac{C_{2}}{\varepsilon}\right)$ & \!\!\!\!\!\!\!  $\mathcal{O}\!\left(n\log\!\frac{C_{2}}{\varepsilon}\!+\!C_{1}\!\sqrt{{nL}/{\varepsilon}}\right)$\!\!\! & \!\!\!\! $\mathcal{O}\!\left(n\log\!\frac{C_{2}}{\varepsilon}\!+\!\frac{\sqrt{n}G}{\sqrt{\mu\varepsilon}}\right)$ & \!\!\!\!\!\!\! $\mathcal{O}\!\left(n\log\frac{C_{2}}{\varepsilon}\!+\!\frac{\sqrt{n}C_{1}G}{\varepsilon}\right)$\!\!\!\! \\
\!\!ASVRG\!\!   & \!\!$\mathcal{O}\!\left((n\!+\!\!\sqrt{n{L}/{\mu}})\log\!\frac{C_{2}}{\varepsilon}\right)$ &\!\!\!\!\!\!\! $\mathcal{O}\!\left(n\log\!\frac{C_{2}}{\varepsilon}\!+\!C_{1}\!\sqrt{{nL}/{\varepsilon}}\right)$\!\!\! & \!\!\!\! $\mathcal{O}\!\left(n\log\!\frac{C_{2}}{\varepsilon}\!+\!\frac{\sqrt{n}G}{\sqrt{\mu\varepsilon}}\right)$ & \!\!\!\!\!\!\! $\mathcal{O}\!\left(n\log\!\frac{C_{2}}{\varepsilon}\!+\!\frac{\sqrt{n}C_{1}G}{\varepsilon}\right)$\!\!\!\! \\
\hline
\end{tabular}
\end{table*}

\section{Preliminaries}
\label{sect20}
Throughout this paper, we use $\|\cdot\|$ to denote the standard Euclidean norm. $\nabla\!f(x)$ denotes the full gradient of $f(x)$ if it is differentiable, or $\partial\!f(x)$ the subgradient if $f(x)$ is Lipschitz continuous. We mostly focus on the case of Problem~\eqref{equ1} when each component function $f_{i}(x)$ is $L_{i}$-smooth\footnote{In the following, we mainly consider the more general class of Problem~\eqref{equ1}, when every $f_{i}(x)$ can have different degrees of smoothness, rather than the gradients of all component functions having the same Lipschitz constant $L$.}, and $g(x)$ is $\mu$-strongly convex.

\begin{assumption}\label{assum1}
Each convex component function $f_{i}(\cdot)$ is $L_{i}$-smooth, if there exists a constant $L_{i}\!>\!0$ such that for all $x,y\!\in\!\mathbb{R}^{d}$, $\|\nabla f_{i}(x)-\nabla f_{i}(y)\|\leq L_{i}\|x-y\|$.
\end{assumption}

\begin{assumption}\label{assum2}
$g(\cdot)$ is $\mu$-strongly convex ($\mu$-SC), if there exists a constant $\mu\!>\!0$ such that for any $x,y\!\in\!\mathbb{R}^{d}$ and any (sub)gradient $\xi$ (i.e., $\xi\!=\!\nabla\!g(x)$, or $\xi\!\in\!\partial g(x)$) of $g(\cdot)$ at $x$
\begin{displaymath}
g(y)\geq g(x)+\xi^{T}(y-x)+\frac{\mu}{2}\|x-y\|^{2}.
\end{displaymath}
\end{assumption}

For a non-strongly convex function, the above inequality can always be satisfied with $\mu\!=\!0$. As summarized in \citep{zhu:box}, there are mainly four interesting cases of Problem~\eqref{equ1}:
\begin{itemize}
\item Case 1: Each $f_{i}(x)$ is $L_{i}$-smooth and $g(x)$ is $\mu$-SC, e.g., ridge regression, logistic regression, and elastic net regularized logistic regression.
\item Case 2: Each $f_{i}(x)$ is $L_{i}$-smooth and $g(x)$ is non-SC, e.g., Lasso and $\ell_{1}$-norm regularized logistic regression.
\item Case 3: Each $f_{i}(x)$ is non-smooth (but Lipschitz continuous) and $g(x)$ is $\mu$-SC, e.g., linear support vector machine (SVM).
\item Case 4: Each $f_{i}(x)$ is non-smooth (but Lipschitz continuous) and $g(x)$ is non-SC, e.g., $\ell_{1}$-norm SVM.
\end{itemize}

\section{Accelerated Proximal SVRG}
In this section, we propose an accelerated proximal stochastic variance reduced gradient (ASVRG) method with momentum acceleration for solving both strongly convex and non-strongly convex objectives (e.g., Cases 1 and 2). Moreover, ASVRG incorporates a weighted sampling strategy as in \citep{xiao:prox-svrg,zhao:sampling,shamir:wrsgd} to randomly pick $i_{t}$ based on a general distribution $\{p_{i},\ldots,p_{n}\}$ rather than the uniform distribution.

\subsection{Iterate Averaging for Snapshot}
Like SVRG, our algorithms are also divided into $S$ epochs, and each epoch consists of $m$ stochastic updates\footnote{In practice, it was reported that reducing the number of gradient
calculations in early iterations can lead to faster convergence \citep{babanezhad:vrsg,zhu:univr,shang:fsvrg}. Thus we set $m_{s+1}\!\!=\!\min(\lfloor\rho m_{s}\rfloor,m)$ in the early epochs of our algorithms, and fix $m_{1}\!=\!n/4$, and $\rho\!=\!2$ without increasing parameter tuning difficulties.}, where $m$ is usually chosen to be $\Theta(n)$ as in \cite{johnson:svrg}. Within each epoch, a full gradient $\nabla\! f(\widetilde{x}^{s-1})$ is calculated at the snapshot $\widetilde{x}^{s-1}$. Note that we choose $\widetilde{x}^{s-1}$ to be the average of the past $t$ stochastic iterates rather than the last iterate because it has been reported to work better in practice \citep{xiao:prox-svrg,flammarion:average,zhu:univr,liu:sadmm,zhu:Katyusha,shang:vrsgd1}. In particular, one of the effects of the choice, i.e., $\widetilde{x}^{s}\!=\!\frac{1}{m}\!\sum^{m}_{t=1}\!x^{s}_{t}$, is to allow taking larger step sizes, e.g., $1/(3L)$ for ASVRG vs.\ $1/(10L)$ for SVRG.

\subsection{ASVRG in Strongly Convex Case}
We first consider the case of Problem (\ref{equ1}) when each $f_{i}(\cdot)$ is $L_{i}$-smooth, and $g(\cdot)$ is $\mu$-SC. Different from existing accelerated methods such as Katyusha \citep{zhu:Katyusha}, we propose a much simpler accelerated stochastic algorithm with momentum, as outlined in Algorithm~\ref{alg1}. Compared with the initialization of $x^{s}_{0}\!=\!y^{s}_{0}\!=\!\widetilde{x}^{s-1}$ (i.e., Option I in Algorithm~\ref{alg1}), the choices of $x^{s+1}_{0}\!=(1\!-\!\omega)\widetilde{x}^{s}+\omega y^{s}_{m_{s}}$ and $y^{s+1}_{0}\!=\!y^{s}_{m_{s}}$ (i.e., Option II) also work well in practice.

\begin{algorithm}[t]
\caption{ASVRG for strongly convex objectives}
\label{alg1}
\renewcommand{\algorithmicrequire}{\textbf{Input:}}
\renewcommand{\algorithmicensure}{\textbf{Initialize:}}
\renewcommand{\algorithmicoutput}{\textbf{Output:}}
\begin{algorithmic}[1]
\REQUIRE The number of epochs $S$, the number of iterations $m$ per epoch, and the step size $\eta$.\\
\ENSURE $\widetilde{x}^{0}\!=x^{0}_{0}=y^{0}_{0}$, \,$\omega$, \,$m_{1}$, \,$\rho\geq1$, and the probability $P=[p_{1},\ldots,p_{n}]$.\\
\FOR{$s=1,2,\ldots,S$}
\STATE {$\widetilde{\mu}=\frac{1}{n}\!\sum^{n}_{i=1}\!\nabla\!f_{i}(\widetilde{x}^{s-1})$;}
\STATE {Option I:\, $x^{s}_{0}=y^{s}_{0}=\widetilde{x}^{s-1}$;}
\FOR{$t=1,2,\ldots,m_{s}$}
\STATE {Pick $i_{t}$ from $\{1,\ldots,n\}$ randomly based on $P$;}
\STATE {$\widetilde{\nabla}\! f_{i_{t}}(x^{s}_{t-1})=[\nabla\! f_{i_{t}}(x^{s}_{t-1})-\nabla\! f_{i_{t}}(\widetilde{x}^{s-1})]/(np_{i_{t}})+\widetilde{\mu}$;}
\STATE {$y^{s}_{t}=\mathop{\arg\min}_{y}\left\{\langle \widetilde{\nabla}\! f_{i_{t}}(x^{s}_{t-1}),\,y-y^{s}_{t-1}\rangle+\frac{\omega}{2\eta}\|y-y^{s}_{t-1}\|^{2}+g(y)\right\}$;}
\STATE {$x^{s}_{t}=\widetilde{x}^{s-1}+\omega(y^{s}_{t}-\widetilde{x}^{s-1})$;}
\ENDFOR
\STATE {$\widetilde{x}^{s}\!=\frac{1}{m_{s}}\!\sum^{m_{s}}_{t=1}\!x^{s}_{t}$,\, $m_{s+1}\!=\min(\lfloor\rho m_{s}\rfloor,m)$;}
\STATE {Option II:\, $x^{s+1}_{0}\!=(1\!-\!\omega)\widetilde{x}^{s}+\omega y^{s}_{m_{s}}$, and\, $y^{s+1}_{0}\!=y^{s}_{m_{s}}$;}
\ENDFOR
\OUTPUT {$\widetilde{x}^{S}$.}
\end{algorithmic}
\end{algorithm}

\subsubsection{Momentum Acceleration}
The update rule of $y$ in our proximal stochastic gradient method is formulated as follows:
\begin{equation}\label{equ5}
\begin{split}
y^{s}_{t}=\mathop{\arg\min}_{y\in\mathbb{R}^{d}}\left\{\langle \widetilde{\nabla}\! f_{i_{t}}(x^{s}_{t-1}),\,y-y^{s}_{t-1}\rangle\!+\!\frac{\omega}{2\eta}\|y-y^{s}_{t-1}\|^{2}+g(y)\right\},
\end{split}
\end{equation}
where $\omega\!\in\![0,1]$ is the momentum parameter. Note that the gradient estimator $\widetilde{\nabla}\! f_{i_{t}}$ used in this paper is the SVRG estimator in (\ref{equ3}). Besides, the algorithms and convergence results of this paper can be generalized to the SAGA estimator in \citep{defazio:saga}. When $g(x)\!\equiv\!0$, the proximal update rule in~\eqref{equ5} degenerates to $y^{s}_{t}\!=y^{s}_{t-1}\!-\!(\eta/{\omega})\widetilde{\nabla}\!f_{i_{t}}\!(x^{s}_{t-1})$.

Inspired by the Nesterov's momentum in \citep{nesterov:fast,nesterov:co,nitanda:svrg,shang:vrsgd} and Katyusha momentum in \citep{zhu:Katyusha}, we design a update rule for $x$ as follows:
\begin{equation}\label{equ9}
x^{s}_{t}=\widetilde{x}^{s-1}+\omega(y^{s}_{t}-\widetilde{x}^{s-1}).
\end{equation}
The second term on the right-hand side of \eqref{equ9} is the proposed momentum similar to the Katyusha momentum in \citep{zhu:Katyusha}. It is clear that there is only one momentum parameter $\omega$ in our algorithm, compared with the two parameters $\omega_{1}$ and $\omega_{2}$ in Katyusha\footnote{Although Acc-Prox-SVRG~\citep{nitanda:svrg} also has a momentum parameter, its oracle complexity is no faster than SVRG when the size of mini-batch is less than $n/2$, as discussed in \citep{zhu:Katyusha}.}.

The per-iteration complexity of ASVRG is dominated by the computation of $\nabla\! f_{i_{t}}\!(x^{s}_{t-1})$ and $\nabla\! f_{i_{t}}\!(\widetilde{x}^{s-1})$,\footnote{For some regularized ERM problems, we can save the intermediate gradients $\nabla\! f_{i_{t}}\!(\widetilde{x}^{s-1})$ in the computation of $\nabla\!f(\widetilde{x}^{s-1})$, which requires $O(n)$ storage in general as in \citep{defazio:Finito}.} and the proximal update in~\eqref{equ5}, which is as low as that of SVRG \citep{johnson:svrg} and Prox-SVRG \citep{xiao:prox-svrg}. In other words, ASVRG has a much lower per-iteration complexity than most of accelerated stochastic methods \citep{murata:dual,zhu:Katyusha} such as Katyusha \citep{zhu:Katyusha}, which has one more proximal update for $z$ in general.

\subsubsection{Momentum Parameter}
Next we give a selection scheme for the stochastic momentum parameter $\omega$. With the given $\eta$, $\omega$ can be a constant, and must satisfy the inequality: $0\!<\!\omega\!\leq\! 1\!-\!\frac{\widetilde{L}\eta}{1-\widetilde{L}\eta}$, where $\widetilde{L}\!=\!\max_{j}L_{j}/(np_{j})$. As shown in Theorem~\ref{theo1} below, it is desirable to have a small convergence factor $\rho$. The following proposition obtains the optimal $\omega_{\star}$, which yields the smallest $\rho$ value.

\begin{proposition}
\label{prop1}
Given a suitable learning rate $\eta$, the optimal parameter $\omega_{\star}$ is $\omega_{\star}\!=\!m\mu\eta/2$.
\end{proposition}

In fact, we can fix $\omega$ to a constant, e.g., $0.9$, which works well in practice as in~\citep{ruder:sgd}. When $\omega\!=\!1$ and $g(x)$ is smooth, Algorithm~\ref{alg1} degenerates to Algorithm 3 in the supplementary material, which is almost identical to SVRG \citep{johnson:svrg}, and the only differences between them are the choice of $\widetilde{x}^{s}$ and the initialization of $x^{s}_{0}$.

\begin{algorithm}[t]
\caption{ASVRG for non-strongly convex objectives}
\label{alg2}
\renewcommand{\algorithmicrequire}{\textbf{Input:}}
\renewcommand{\algorithmicensure}{\textbf{Initialize:}}
\renewcommand{\algorithmicoutput}{\textbf{Output:}}
\begin{algorithmic}[1]
\REQUIRE The number of epochs $S$, the number of iterations $m$ per epoch, and the step size $\eta$.\\
\ENSURE $\widetilde{x}^{0}\!=\widetilde{y}^{0}$, $\omega_{0}=1\!-\!\frac{\widetilde{L}\eta}{1-\widetilde{L}\eta}$, $m_{1}$, $\rho\geq1$, and $P=[p_{1},\ldots,p_{n}]$.\\
\FOR{$s=1,2,\ldots,S$}
\STATE {$\widetilde{\mu}=\frac{1}{n}\!\sum^{n}_{i=1}\!\!\nabla\!f_{i}(\widetilde{x}^{s-1})$,\, $x^{s}_{0}=(1\!-\!\omega_{s-1})\widetilde{x}^{s-1}+\omega_{s-1}\widetilde{y}^{s-1}$,\, $y^{s}_{0}=\widetilde{y}^{s-1}$;}
\FOR{$t=1,2,\ldots,m_{s}$}
\STATE {Pick $i_{t}$ from $\{1,\ldots,n\}$ randomly based on $P$;}
\STATE {$\widetilde{\nabla}\! f_{i_{t}}(x^{s}_{t-1})=[\nabla\! f_{i_{t}}(x^{s}_{t-1})-\nabla\! f_{i_{t}}(\widetilde{x}^{s-1})]/(np_{i_{t}})+\widetilde{\mu}$;}
\STATE {$y^{s}_{t}=\mathop{\arg\min}_{y}\left\{\langle \widetilde{\nabla}\! f_{i_{t}}(x^{s}_{t-1}),\,y-y^{s}_{t-1}\rangle+\frac{\omega}{2\eta}\|y-y^{s}_{t-1}\|^{2}+g(y)\right\}$;}
\STATE {$x^{s}_{t}=\widetilde{x}^{s-1}+\omega_{s-1}(y^{s}_{t}-\widetilde{x}^{s-1})$;}
\ENDFOR
\STATE {$\widetilde{x}^{s}\!=\frac{1}{m_{s}}\!\sum^{m_{s}}_{t=1}\!x^{s}_{t}$,\, $\widetilde{y}^{s}\!=y^{s}_{m_{s}}$,\, $\omega_{s}\!=\frac{\sqrt{\omega^{4}_{s-1}+4\omega^{2}_{s-1}}-\;\!\omega^{2}_{s-1}}{2}$,\, $m_{s+1}\!=\min(\lfloor\rho m_{s}\rfloor,m)$;}
\ENDFOR
\OUTPUT {$\widetilde{x}^{S}$.}
\end{algorithmic}
\end{algorithm}

\subsection{ASVRG in Non-Strongly Convex Case}
We also develop an efficient algorithm for solving non-SC problems, as outlined in Algorithm~\ref{alg2}. The main difference between Algorithms~\ref{alg1} and~\ref{alg2} is the setting of the momentum parameter. That is, the momentum parameter $\omega_{s}$ in Algorithm~\ref{alg2} is decreasing, while that of Algorithm~\ref{alg1} can be a constant. Different from Algorithm~\ref{alg1}, $\omega_{s}$ in Algorithm~\ref{alg2} needs to satisfy the following inequalities:
\begin{equation}\label{equ11}
\frac{1-\omega_{s}}{\omega^{2}_{s}}\leq \frac{1}{\omega^{2}_{s-\!1}}\;\,\textup{and}\;\, 0\leq\omega_{s}\leq 1-\frac{\widetilde{L}\eta}{1-\widetilde{L}\eta}.
\end{equation}
It is clear that the condition (\ref{equ11}) allows the stochastic momentum parameter to decrease, but not too fast, similar to the requirement on the step-size $\eta_{t}$ in classical SGD. Unlike deterministic acceleration methods, where $\omega_{s}$ is only required to satisfy the first inequality in~(\ref{equ11}), the momentum parameter $\omega_{s}$ in Algorithm~\ref{alg2} must satisfy both inequalities. Inspired by the momentum acceleration techniques in \citep{tseng:sco,su:nag} for deterministic optimization, the update rule for $\omega_{s}$ is defined as follows: $\omega_{0}\!=\!1\!-\!{\widetilde{L}\eta}/({1\!-\!\widetilde{L}\eta})$, and for any $s\!>\!1$, $\omega_{s}\!=\!\frac{\sqrt{\omega^{4}_{s-1}+4\omega^{2}_{s-1}}-\omega^{2}_{s-1}}{2}$.

\section{Convergence Analysis}
\label{sect40}
In this section, we provide the convergence analysis of ASVRG for both strongly convex and non-strongly convex objectives. We first give the following key intermediate result (the proofs to all theoretical results in this paper are given in the supplementary material).

\begin{lemma}
\label{lemm1}
Suppose Assumption~\ref{assum1} holds. Let $x^{\star}$ be an optimal solution of Problem~\eqref{equ1}, and $\{(x^{s}_{t},y^{s}_{t})\}$ be the sequence generated by Algorithms~\ref{alg1} and \ref{alg2} with $\omega_{s}\leq 1\!-\!\frac{\widetilde{L}\eta}{1-\widetilde{L}\eta}$\footnote{Note that the momentum parameter $\omega_{s}$ in Algorithm~\ref{alg1} is a constant, that is, $\omega_{s}\equiv\omega$. In addition, if the length of the early epochs is not sufficiently large, the epochs can be viewed as an initialization step}. Then for all $s\!=\!1,\ldots,S$,
\begin{equation*}
\begin{split}
\mathbb{E}\!\left[F(\widetilde{x}^{s})\!-\!F(x^{\star})\right]\leq(1\!-\!\omega_{s-1})\mathbb{E}\!\left[F(\widetilde{x}^{s-1})-F(x^{\star})\right]\!+\frac{\omega^{2}_{s-1}}{2m\eta}\mathbb{E}\!\left[\|x^{\star}\!-y^{s}_{0}\|^2\!-\!\|x^{\star}\!-y^{s}_{m}\|^2\right]\!.
\end{split}
\end{equation*}
\end{lemma}

\subsection{Analysis for Strongly Convex Objectives}
\label{sect41}
For strongly convex objectives, our first main result is the following theorem, which gives the convergence rate and oracle complexity of Algorithm~\ref{alg1}.

\begin{figure}[t]
\centering
\subfigure[Corollary~\ref{coro1}]{\includegraphics[width=0.486\columnwidth]{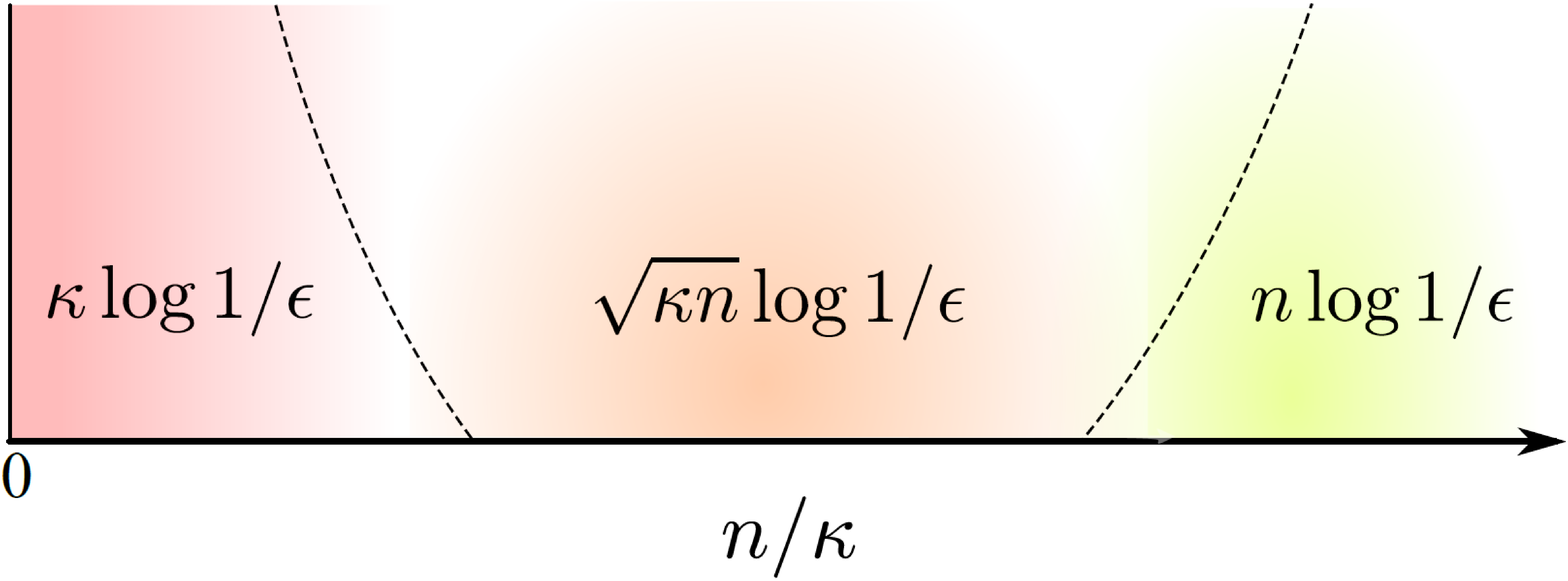}}\:
\subfigure[Corollary~\ref{coro21}]{\includegraphics[width=0.486\columnwidth]{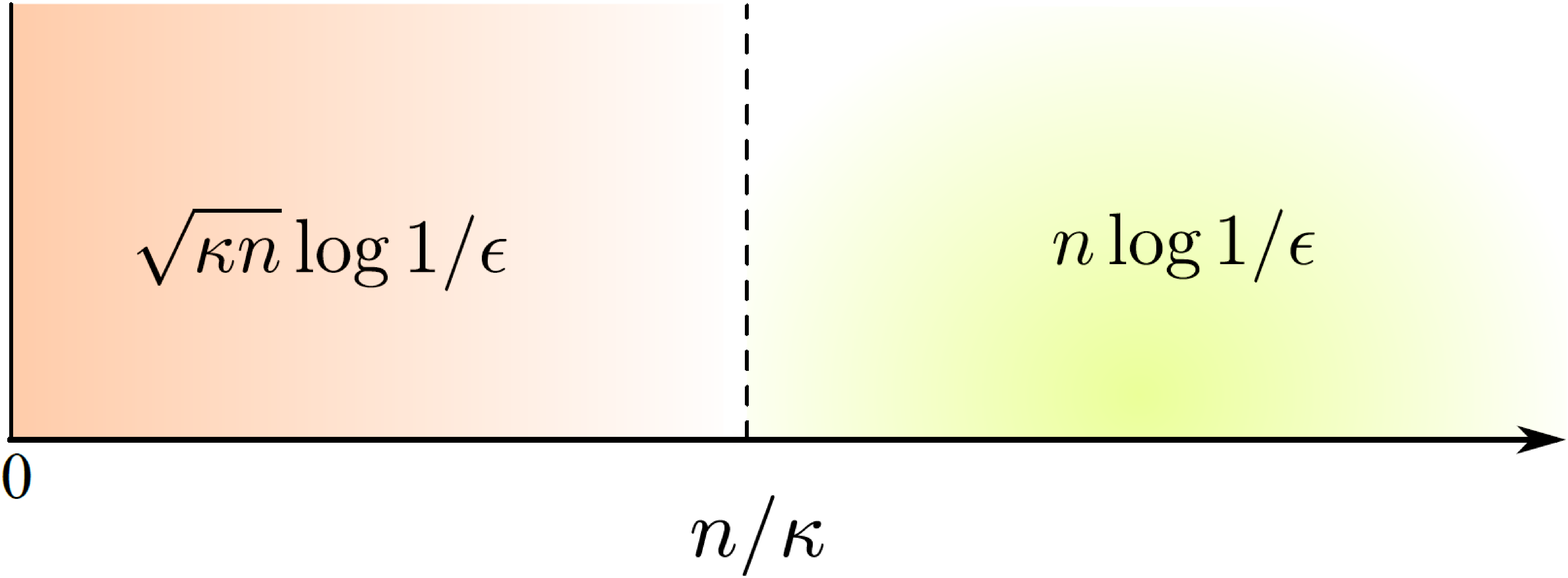}}
\vspace{-3mm}
\caption{Comparison of the oracle complexities of Algorithm~\ref{alg1} with Option I and Option II.}
\label{figs01}
\end{figure}

\begin{theorem}\label{theo1}
Suppose Assumptions~\ref{assum1} and \ref{assum2} hold, and given the same notation as in Lemma~\ref{lemm1}, and $m$ is sufficiently large so that
\begin{equation*}
\rho:=\,1-\omega+\frac{\omega^{2}}{m\mu\eta}< 1.
\end{equation*}
Then Algorithm~\ref{alg1} with Option I has the following geometric convergence in expectation:
\begin{equation*}
\mathbb{E}\!\left[F(\widetilde{x}^{s})-F(x^{\star})\right]\leq\,\rho^{s}\!\left[F(\widetilde{x}^{0})-F(x^{\star})\right].
\end{equation*}
\end{theorem}
Theorem~\ref{theo1} shows that Algorithm~\ref{alg1} with Option I achieves linear convergence for strongly convex problems. We can easily obtain a similar result for Algorithm~\ref{alg1} with Option II. The following results give the oracle complexities of Algorithm~\ref{alg1} with Option I or Option II, as shown in Figure~\ref{figs01}, where $\kappa\!=\!\widetilde{L}/\mu$ is the condition number.

\begin{corollary}\label{coro1}
The oracle complexity of Algorithm~\ref{alg1} with Option I to achieve an $\varepsilon$-suboptimal solution (i.e., $\mathbb{E}\!\left[F(\widetilde{x}^{s})\right]-F(x^{\star})\leq \varepsilon$) is
\[                                                                                                                                                                                                                                                       \begin{cases}                                                                                            \mathcal{O}\big(\sqrt{{n\widetilde{L}}/{\mu}}\log{\frac{F(\tilde{x}^{0})-F(x^{\star})}{\varepsilon}}\big), & \;\;\;\,\text{if }\, {m\mu}/{\widetilde{L}}\!\in\![0.68623,145.72],\\                                     \mathcal{O}\big((n+{\widetilde{L}}/{\mu})\log{\frac{F(\tilde{x}^{0})-F(x^{\star})}{\varepsilon}}\big), &\quad\text{otherwise}.                                                                                                                                                                                                      \end{cases}
\]                                                                                                                                                                                                                                                            \end{corollary}

\begin{corollary}\label{coro21}
The oracle complexity of Algorithm~\ref{alg1} with Option II and restarts every $\mathcal{S}=2\cdot\big(\frac{1 - \omega}{\omega} + \frac{\omega}{\eta m \mu}\big)$ epochs\footnote{For each restart, the new initial point is set to $\widetilde{x}^{0}\!=\!\frac{1}{\mathcal{S}}\!\sum^{\mathcal{S}}_{s=1}\!\widetilde{x}^{s}$. If we choose the snapshot to be the weighted average as in \citep{zhu:Katyusha} rather than the uniform average, our algorithm without restarts can also achieve the tightest possible result. Note that $\mathcal{O}(n\!+\!\!\sqrt{nL/\mu})$ is always less than $\mathcal{O}(n\!+\!L/\mu)$.} to achieve an $\varepsilon$-suboptimal solution (i.e., $\mathbb{E}\!\left[F(\widetilde{x}^{s})\right]-F(x^{\star})\leq \varepsilon$) is
\[
\mathcal{O}\big((n\!+\!\sqrt{{n\widetilde{L}}/{\mu}})\log{\frac{F(\tilde{x}^{0}) - F(x^{\star})}{\varepsilon}}\big).
\]
\end{corollary}

For the most commonly used uniform random sampling (i.e., sampling probabilities $p_{i}\!=\!1/n$ for all $i\!=\!1,\ldots,n$), the oracle complexity of ASVRG becomes $\mathcal{O}((n\!+\!\!\sqrt{nL/\mu})\log({1}/{\varepsilon}))$, which is identical to that of Katyusha \citep{zhu:Katyusha} and Point-SAGA \citep{defazio:sagab}, and better than those of non-accelerated methods (e.g., $\mathcal{O}\!\left((n\!+\!L/\mu)\log({1}/{\varepsilon})\right)$ of SVRG).

\begin{remark}
For uniform random sampling, and recalling $L\!=\!L_{\max}\!=\!\max_{j}L_{j}$, the above oracle bound is rewritten as: $\mathcal{O}((n\!+\!\sqrt{nL_{\max}/\mu})\log({1}/{\varepsilon}))$. As each $f_{i}(\cdot)$ generally has different degrees of smoothness $L_{i}$, picking the random index $i_{t}$ from a non-uniform distribution is a much better choice than simple uniform random sampling \citep{zhao:sampling,needell:sgd}. For instance, the sampling probabilities for all $f_{i}(\cdot)$ are proportional to their Lipschitz constants, i.e., $p_{i}\!=\!L_{i}/\sum^{n}_{j=1}\!L_{j}$. In this case, the oracle complexity becomes $\mathcal{O}((n\!+\!\sqrt{nL_{\textup{avg}}/\mu})\log({1}/{\varepsilon}))$, where $L_{\textup{avg}}\!=\!\frac{1}{n}\!\sum^{n}_{j=1}\!L_{j}\!\leq\!L_{\max}$. In other words, the statement of Corollary~\ref{coro21} can be revised by simply replacing $\widetilde{L}$ with $L_{\textup{avg}}$. And the proof only needs some minor changes accordingly. In fact, all statements in this section can be revised by replacing $\widetilde{L}$ with $L_{\textup{avg}}$.
\end{remark}

\subsection{Analysis for Non-Strongly Convex Objectives}
\label{sect42}
For non-strongly convex objectives, our second main result is the following theorem, which gives the convergence rate and oracle complexity of Algorithm~\ref{alg2}.

\begin{theorem}\label{theo2}
Suppose Assumption~\ref{assum1} holds. Then the following result holds,
\begin{equation*}
\begin{split}
\mathbb{E}\!\left[F(\widetilde{x}^{S})-F(x^{\star})\right]
\leq\frac{4(\alpha\!-\!1)[F(\widetilde{x}^{0})-F(x^{\star})]}{(\alpha-2)^{2}(S+1)^{2}}+\frac{2\|x^{\star}-\widetilde{x}^{0}\|^2}{\eta m(S+1)^{2}},
\end{split}
\end{equation*}
where $\alpha\!=\!{1}/{(\widetilde{L}\eta)}\!>\!2$. Furthermore, choosing $m\!=\!\Theta(n)$, Algorithm~\ref{alg2} achieves an $\varepsilon$-suboptimal solution, i.e., $\mathbb{E}[F(\widetilde{x}^{S})]-F(x^{\star})\leq \varepsilon$ using at most $\mathcal{O}(\frac{n\sqrt{F(\widetilde{x}^{0})-F(x^{\star})}}{\sqrt{\varepsilon}}+\frac{\sqrt{n\widetilde{L}}\|\widetilde{x}^{0}-x^{\star}\|}{\sqrt{\varepsilon}})$ iterations.
\end{theorem}
One can see that the oracle complexity of ASVRG is consistent with the best known result in \citep{hien:asmd,zhu:Katyusha}, and all the methods attain the optimal convergence rate $\mathcal{O}(1/t^2)$. Moreover, we can use the adaptive regularization technique in~\citep{zhu:box} to the original non-SC problem, and achieve a SC objective $F(x)\!+\!\frac{\sigma_{s}}{2}\|x-\widetilde{x}^{0}\|$ with a decreasing value $\sigma_{s}$, e.g., $\sigma_{s}\!=\!\sigma_{s-1}/2$ in~\citep{zhu:box}. Then we have the following result.

\begin{corollary}\label{coro2}
Suppose Assumption~\ref{assum1} holds, and $g(\cdot)$ is non-SC. By applying the adaptive regularization technique in~\citep{zhu:box} for Algorithm~\ref{alg1}, then we obtain an $\varepsilon$-suboptimal solution using at most the following oracle complexity:
\begin{equation*}
\mathcal{O}\!\left(n\log\frac{F(\widetilde{x}^{0})-F(x^{\star})}{\varepsilon}+\frac{\sqrt{n\widetilde{L}}\|\widetilde{x}^{0}-x^{\star}\|}{\sqrt{\varepsilon}}\right).
\end{equation*}
\end{corollary}
Corollary~\ref{coro2} implies that ASVRG has a low oracle complexity (i.e., $\mathcal{O}(n\log(1/\varepsilon)\!+\!C_{1}\sqrt{n\widetilde{L}/\varepsilon})$), which is the same as that in \citep{zhu:Katyusha}. Both ASVRG and Katyusha have a much faster rate than SAGA~\citep{defazio:saga}, whose oracle complexity is $\mathcal{O}((n\!+\!L)/\varepsilon)$.

Although ASVRG is much simpler than Katyusha, all the theoretical results show that ASVRG achieves the same convergence rates and oracle complexities as Katyusha for both SC and non-SC cases. Similar to \citep{babanezhad:vrsg,zhu:univr}, we can reduce the number of gradient calculations in early iterations to further speed up ASVRG in practice.

\section{Extensions of ASVRG}
In this section, we first extend ASVRG to the mini-batch setting. Then we extend Algorithm~\ref{alg1} and its convergence results to the non-smooth setting (e.g., the problems in Cases 3 and 4).

\subsection{Mini-Batch}
In this part, we extend ASVRG and its convergence results to the mini-batch setting. Suppose that the mini-batch size is $b$, the stochastic gradient estimator with variance reduction becomes
\begin{equation*}
\widetilde{\nabla}\! f_{I_{t}}\!(x^{s}_{t-1})=\frac{1}{b}\sum_{i\in I_{t}}\!\frac{1}{np_{i}}\!\left[\nabla\!f_{i}(x^{s}_{t-1})-\nabla\! f_{i}(\widetilde{x}^{s-1})\right]+{\nabla}\! f(\widetilde{x}^{s-1}),
\end{equation*}
where $I_{t}\!\subset\![n]$ is a mini-batch of size $b$. Consequently, the momentum parameters $\omega$ and $\omega_{s}$ must satisfy $\omega\!\leq\! 1\!-\!\frac{\tau(b)\widetilde{L}\eta}{1-\widetilde{L}\eta}$ and $\omega_{s}\!\leq\! 1\!-\!\frac{\tau(b)\widetilde{L}\eta}{1-\widetilde{L}\eta}$ for SC and non-SC cases, respectively, where $\tau(b)\!=\!(n\!-\!b)/[b(n\!-\!1)]$. Moreover, the upper bound on the variance of $\widetilde{\nabla}\!f_{i_{t}}(x^{s}_{t-\!1})$ can be extended to the mini-batch setting as follows.

\begin{lemma}\label{lemm3.1}
\begin{displaymath}
\begin{split}
\mathbb{E}\!\left[\left\|\widetilde{\nabla}\! f_{I_{t}}\!(x^{s}_{t-1})-\nabla\! f(x^{s}_{t-1})\right\|^{2}\right]
\leq\frac{2\widetilde{L}(n\!-\!b)}{b(n\!-\!1)}\!\!\left[f(\widetilde{x}^{s-1})\!-\!f(x^{s}_{t-1})\!+\!\nabla\! f(x^{s}_{t-1})^{T}\!(x^{s}_{t-1}\!-\!\widetilde{x}^{s-1})\right]\!.
\end{split}
\end{displaymath}
\end{lemma}

In the SC case, the convergence result of the mini-batch variant\footnote{Note that in the mini-batch setting, the number of stochastic iterations of the inner loop in Algorithms~\ref{alg1} and \ref{alg2} is reduced from $m$ to $\lfloor m/b\rfloor$.} of ASVRG is identical to Theorem~\ref{theo1} and Corollary~\ref{coro21}. For the non-SC case, we set the initial parameter $\omega_{0}\!=\!1\!-\!\frac{\tau(b)\widetilde{L}\eta}{1-\widetilde{L}\eta}$. Then Theorem~\ref{theo2} can be extended to the mini-batch setting as follows.

\begin{theorem}\label{theo3}
Suppose Assumption~\ref{assum1} holds, and given the same notation as in Theorem~\ref{theo2} and $\omega_{0}\!=\!1-\frac{\tau(b)\widetilde{L}\eta}{1-\widetilde{L}\eta}$, then the following  inequality holds
\begin{equation}\label{equ15}
\begin{split}
\mathbb{E}\!\left[F(\widetilde{x}^{s})-F(x^{\star})\right]
\leq\frac{4(\alpha\!-\!1)\tau(b)[F(\widetilde{x}^{0})\!-\!F(x^{\star})]}{[\alpha-1-\tau(b)]^{2}(s+1)^{2}}+\frac{2\widetilde{L}\alpha\|x^{\star}\!-\!\widetilde{x}^{0}\|^2}{m(s+1)^{2}}.
\end{split}
\end{equation}
\end{theorem}

\begin{remark}
For the special case of $b\!=\!1$, we have $\tau(1)\!=\!1$, and then Theorem ~\ref{theo3} degenerates to Theorem ~\ref{theo2}. When $b\!=\!n$ (i.e., the batch setting), then $\tau(n)\!=\!0$, and the first term on the right-hand side of \eqref{equ15} diminishes. Then our algorithm degenerates to an accelerated deterministic method with the optimal convergence rate of $\mathcal{O}(1/t^{2})$, where $t$ is the number of iterations.
\end{remark}

\subsection{Non-Smooth Settings}
In addition to the application of the regularization reduction technique in~\citep{zhu:box} for a class of smooth and non-SC problems (i.e., Case 2 of Problem~\eqref{equ1}), as shown in Section~\ref{sect42}, ASVRG can also be extended to solve the problems in both Cases 3 and 4, when each component of $f$ is $G$-Lipschitz continuous, which is defined as follows.
\begin{definition}
The function $f_{i}(x)$ is $G$-Lipschitz continuous if there exists a constant $G$ such that, for any $x,y\!\in\!\mathbb{R}^{d}$, $|f_{i}(x)-f_{i}(y)|\leq G\|x-y\|$.
\end{definition}

The key technique is using a proximal operator to obtain gradients of the $\delta_{s}$-Moreau envelope of a non-smooth function $f_{i}(\cdot)$, defined as

\begin{equation}
f^{\delta_{s}}_{i}(x)=\inf_{y\in\mathbb{R}^{d}} f_{i}(y)+\frac{\delta_{s}}{2}\|x-y\|^{2},
\end{equation}
where $\delta_{s}$ is an increasing parameter as in \citep{zhu:box}. That is, we use the proximal operator to smoothen each component function, and optimize the new, smooth function which approximates the original problem. This technique has been used in Katyusha~\citep{zhu:Katyusha} and accelerated SDCA~\citep{shalev-Shwartz:acc-sdca} for non-smooth objectives.

\begin{property}[\cite{nesterov:smooth,bauschke:book,orabona:smooth}]
Let each $f_{i}(x)$ be convex and $G$-Lipschitz continuous. For any $\delta>0$, the following results hold:\\
(a) $f^{\delta}_{i}(x)$ is $\delta$-smooth;\\
(b) $\nabla(f^{\delta}_{i})(x)=\delta\!\left(x-\textup{prox}^{\,f_{i}}_{\delta}(x)\right)$;\\
(c) $f^{\delta}_{i}(x)\leq f_{i}(x)\leq f^{\delta}_{i}(x)+\frac{G^{2}}{2\delta}$.
\end{property}

By using the similar techniques in~\citep{zhu:box}, we can apply ASVRG to solve the smooth problem $F^{\delta_{s}}(x)\!:=\!\frac{1}{n}\!\sum^{n}_{i=1}\!f^{\delta_{s}}_{i}(x)\!+\!g(x)$. It is easy to verify that ASVRG satisfies the homogenous objective decrease (HOOD) property in~\citep{zhu:box} (see \citep{zhu:box} for the detail of HOOD), as shown below.

\begin{corollary}\label{coro3}
Algorithm~\ref{alg1} used to solve the problem in Case 1 satisfies the HOOD property with at most $\mathcal{O}(n\!+\!\sqrt{n\widetilde{L}/\mu})$ iterations. That is, for every starting point $\widetilde{x}^{0}$, Algorithm~\ref{alg1} produces an output $\widetilde{x}^{s}$ satisfying $\mathbb{E}\!\left[F(\widetilde{x}^{s})\right]\!-\!F(x^{\star})\!\leq\! [F(\widetilde{x}^{0})\!-\!F(x^{\star})]/4$ in at most $\mathcal{O}(n\!+\!\sqrt{n\widetilde{L}/\mu})$ iterations.
\end{corollary}

In the following, we extend the result in Theorem~\ref{theo1} to the non-smooth setting as follows.
\begin{corollary}\label{coro4}
Let $f_{i}(x)$ be $G$-Lpischitz continuous, and $g(x)$ be $\mu$-strongly convex. By applying the adaptive smooth technique in~\citep{zhu:box} on ASVRG, we obtain an $\varepsilon$-suboptimal solution using at most the following oracle complexity:
\begin{equation*}
\mathcal{O}\!\left(n\log\frac{F(\widetilde{x}^{0})-F(x^{\star})}{\varepsilon}+\frac{\sqrt{n}G}{\sqrt{\mu\varepsilon}}\right).
\end{equation*}
\end{corollary}

\begin{corollary}\label{coro5}
Let $f_{i}(x)$ be $G$-Lpischitz continuous and $g(x)$ be not necessarily strongly convex. By applying both the adaptive regularization and smooth techniques in~\citep{zhu:box} on ASVRG, we obtain an $\varepsilon$-suboptimal solution using at most the following oracle complexity:
\begin{equation*}
\mathcal{O}\!\left(n\log\frac{F(\widetilde{x}^{0})-F(x^{\star})}{\varepsilon}+\frac{\sqrt{n}G\|\widetilde{x}^{0}-x^{\star}\|}{\varepsilon}\right).
\end{equation*}
\end{corollary}

From Corollaries~\ref{coro4} and \ref{coro5}, one can see that ASVRG converges to an $\varepsilon$-accurate solution for Case 3 of Problem~\eqref{equ1} in $\mathcal{O}\!\left(n\log\frac{C_{2}}{\varepsilon}\!+\!\!\frac{\sqrt{n}G}{\sqrt{\mu\varepsilon}}\right)$ iterations and for Case 4 in $\mathcal{O}\!\left(n\log\frac{C_{2}}{\varepsilon}\!+\!\!\frac{\sqrt{n}C_{1}G}{\varepsilon}\right)$ iterations. That is, ASVRG achieves the same low oracle complexities as the accelerated stochastic variance reduction method, Katyusha, for the two classes of non-smooth problems (i.e., Cases 3 and 4 of Problem~\eqref{equ1}).


\begin{figure}[t]
\centering
\subfigure[RCV1, \,$\lambda=10^{-6}$]{\includegraphics[width=0.469\columnwidth]{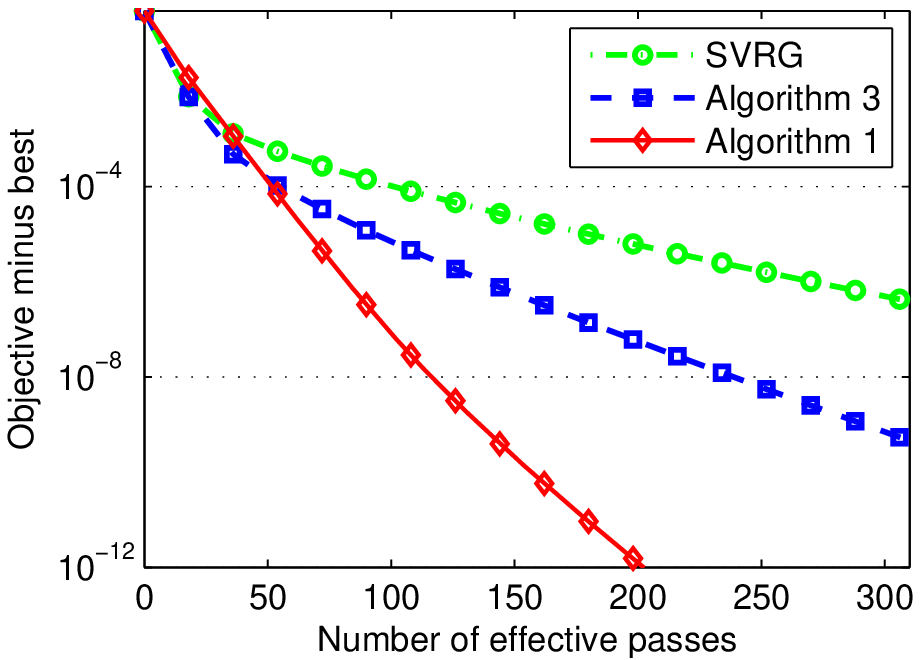}}\;\,\;\:
\subfigure[Covtype, \,$\lambda=10^{-7}$]{\includegraphics[width=0.469\columnwidth]{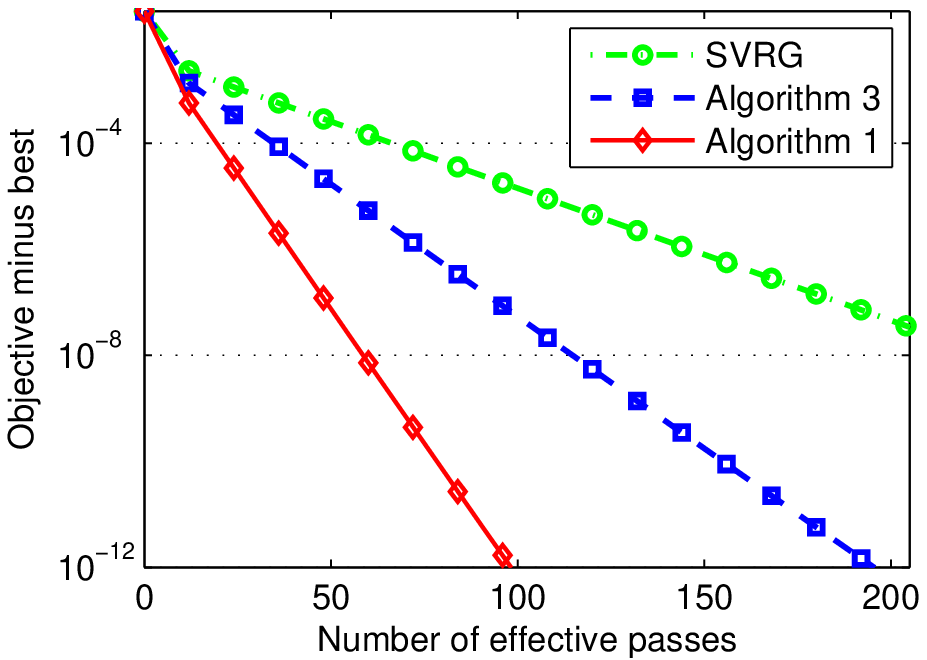}}
\vspace{-3mm}
\caption{Comparison of SVRG~\protect\cite{johnson:svrg}, ASVRG without momentum (i.e., Algorithm 3), and ASVRG (i.e., Algorithm~\ref{alg1}) for ridge regression.}
\label{figs02}
\end{figure}

\section{Experiments}
In this section, we evaluate the performance of ASVRG, and all the experiments were performed on a PC with an Intel i5-2400 CPU and 16GB RAM. We used the two publicly available data sets in our experiments: Covtype and RCV1, which can be downloaded from the LIBSVM Data website\footnote{\url{https://www.csie.ntu.edu.tw/~cjlin/libsvm/}}.

\subsection{Effectiveness of Our Momentum}
Figure~\ref{figs02} shows the performance of ASVRG with $\theta_{s}\!\equiv\!1$ (i.e., Algorithm 3) and ASVRG in order to illustrate the importance and effectiveness of our momentum. Note that the epoch length is set to $m\!=\!2n$ for the two algorithms (i.e., $m_{1}\!=\!2n$ and $\rho\!=\!1$), as well as SVRG \citep{johnson:svrg}. It is clear that Algorithm 3 is without our momentum acceleration technique. The main difference between Algorithm 3 and SVRG is that the snapshot and starting points of the former are set to the uniform average and last iterate of the previous epoch, respectively, while the two points of the later are the last iterate. The results show that Algorithm 3 outperforms SVRG, suggesting that the iterate average can work better in practice, as discussed in \citep{shang:vrsgd1}. In particular, ASVRG converges significantly faster than ASVRG without momentum and SVRG, meaning that our momentum acceleration technique can accelerate the convergence of ASVRG.

\subsection{Comparison with Stochastic Methods}
For fair comparison, ASVRG and the compared algorithms (including SVRG~\citep{johnson:svrg}, SAGA~\citep{defazio:saga}, Acc-Prox-SVRG~\citep{nitanda:svrg}, Catalyst~\citep{lin:vrsg}, and Katyusha~\citep{zhu:Katyusha}) were implemented in C++ with a Matlab interface. There is only one parameter (i.e., the learning rate) to tune for all these methods except Catalyst and Acc-Prox-SVRG. In particular, we compare their performance in terms of both the number of effective passes over the data and running time (seconds). As in \citep{xiao:prox-svrg}, each feature vector has been normalized so that it has norm $1$.

\begin{figure}[t]
\centering
\includegraphics[width=0.469\columnwidth]{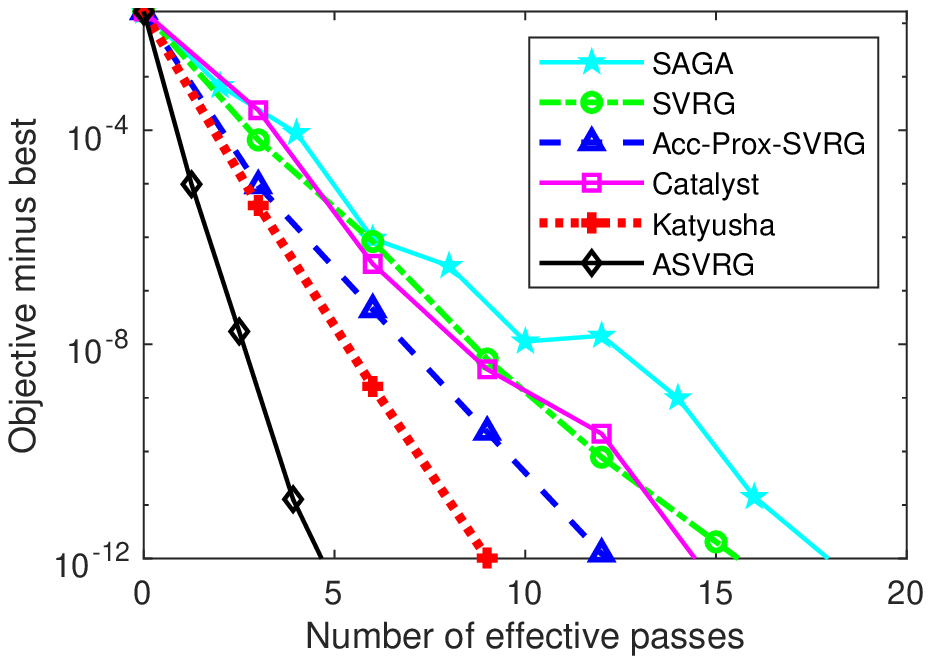}\;\;\,\includegraphics[width=0.469\columnwidth]{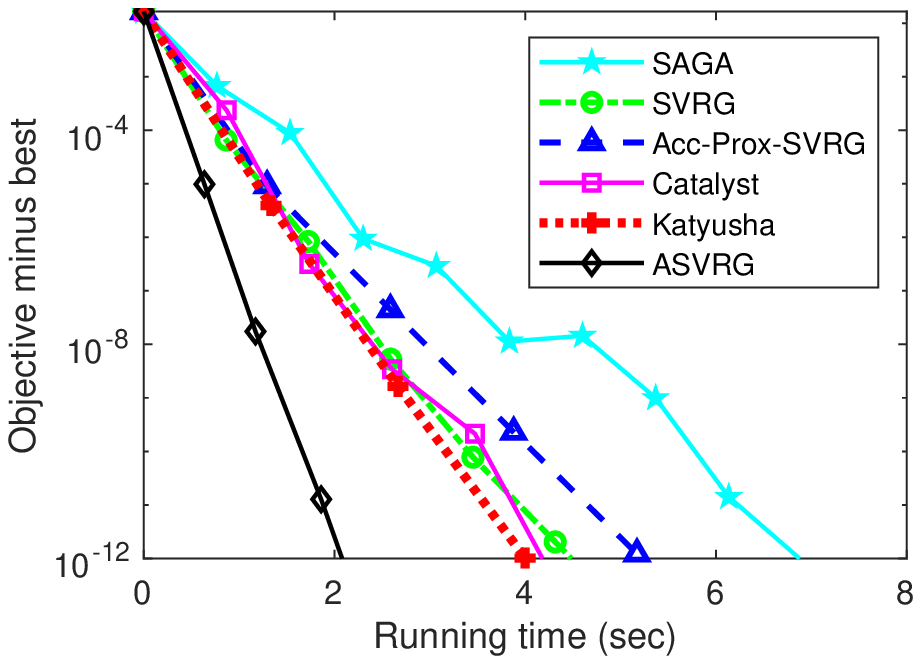}
\vspace{-3mm}
\caption{Comparison of SAGA \citep{defazio:saga}, SVRG \citep{johnson:svrg}, Acc-Prox-SVRG~\citep{nitanda:svrg}, Catalyst \citep{lin:vrsg}, Katyusha \citep{zhu:Katyusha} and ASVRG on Covtype.}
\label{figs03}
\end{figure}

Figure~\ref{figs03} shows how the objective gap (i.e., $F(x^{s})\!-\!F(x^{\star})$) of all these methods decreases on elastic net regularized logistic regression ($\lambda_{1}\!=\!10^{-4}$, $\lambda_{2}\!=\!10^{-5}$) as time goes on. It is clear that ASVRG converges significantly faster than the other methods in terms of both oracle calls and running time, while Catalyst and Katyusha achieve comparable and sometimes even better performance than SVRG and SAGA in terms of the running time (seconds). The main reason is that ASVRG not only takes advantage of the momentum acceleration trick, but also can use much larger step-size (e.g., 1/(3$L$) for ASVRG vs.\ 1/(10$L$) for SVRG). This empirically verifies our theoretical result in Corollary~\ref{coro21} that ASVRG has the same low oracle complexity as Katyusha. ASVRG significantly outperforms Katyusha in terms of running time, which implies that ASVRG has a much lower per-iteration cost than Katyusha.

\section{Conclusions}
We proposed an efficient ASVRG method, which integrates both the momentum acceleration trick and variance reduction technique. We first designed a simple momentum acceleration technique. Then we theoretically analyzed the convergence properties of ASVRG, which show that ASVRG achieves the same low oracle complexities for both SC and non-SC objectives as accelerated methods, e.g., Katyusha~\citep{zhu:Katyusha}. Moreover, we also extended ASVRG and its convergence results to both mini-batch settings and non-smooth settings.

It would be interesting to consider other classes of settings, e.g., the non-Euclidean norm setting. In practice, ASVRG is much simpler than the existing accelerated methods, and usually converges much faster than them, which has been verified in our experiments. Due to its simplicity, it is more friendly to asynchronous parallel and distributed implementation for large-scale machine learning problems~\citep{zhou:fsvrg}, similar to~\citep{reddi:sgd,sra:dsgd,mania:svrg,zhou:fsvrg,lee:dsgd,wang:dsvrg}. One natural open problem is whether the best oracle complexities can be obtained by ASVRG in the asynchronous and distributed settings.

\subsubsection*{Acknowledgments}
We thank the reviewers for their valuable comments. This work was supported in part by Grants (CUHK 14206715 \& 14222816) from the Hong Kong RGC, Project supported the Foundation for Innovative Research Groups of the National Natural Science Foundation of China (No.\ 61621005), the Major Research Plan of the National Natural Science Foundation of China (Nos.\ 91438201 and 91438103), the National Natural Science Foundation of China (Nos.\ 61876220, 61876221, 61836009, U1701267, 61871310, 61573267, 61502369 and 61473215), the Program for Cheung Kong Scholars and Innovative Research Team in University (No.\ IRT\_15R53), the Fund for Foreign Scholars in University Research and Teaching Programs (the 111 Project) (No.\ B07048), and the Science Foundation of Xidian University (No.\ 10251180018).

\bibliography{icml2018}

\setcounter{algorithm}{2}
\setcounter{equation}{9}
\setcounter{property}{1}

~\\
~\\

\appendix
In this supplementary material, we give the detailed proofs for some lemmas, theorems and properties.

\section{}
\subsection*{Appendix A1: Proof of Proposition 1}
\begin{proof}
Using Theorem 1, we have
\begin{equation*}
\rho(\omega)=1-\omega+\frac{\omega^{2}}{\mu m\eta}.
\end{equation*}
Obviously, it is desirable to have a small convergence factor $\rho(\omega)$. So, we minimize $\rho(\omega)$ with given $\eta$. Then we have
\begin{equation*}
\omega_{\star}=m\mu\eta/2\leq 1-\frac{\widetilde{L}\eta}{1-\widetilde{L}\eta},
\end{equation*}
and
\begin{equation*}
\rho(\omega_{\star})=1-\frac{m\mu\eta}{4}>0.
\end{equation*}

The above two inequalities imply that
\begin{equation*}
\eta\leq\frac{1+4c_{1}-\sqrt{1+16c_{1}^2}}{2\widetilde{L}}=\frac{1+4c_{1}-\sqrt{1+16c_{1}^2}}{2c_{1} m\mu} \quad \textrm{and}\quad \eta<\frac{4}{m\mu},
\end{equation*}
where $c_{1}={\widetilde{L}}/{(m\mu)}>0$. This completes the proof.
\end{proof}

\vspace{2mm}

\subsection*{Appendix A2: ASVRG Pseudo-Codes}
We first give the details on Algorithm 1 with $\omega\!=\!1$ for optimizing smooth objective functions such as $\ell_2$-norm regularized logistic regression, as shown in Algorithm~\ref{alg3}, which is almost identical to the regularized SVRG in~\citep{babanezhad:vrsg} and the original SVRG in~\citep{johnson:svrg}. The main differences between Algorithm~\ref{alg3} and the latter two are the initialization of $x^{s}_{0}$ and the choice of the snapshot point $\widetilde{x}^{s}$. Moreover, we can use the doubling-epoch technique in~\citep{mahdavi:sgd,zhu:univr} to further speed up our ASVRG method for both SC and non-SC cases. Besides, all the proposed algorithms can be extended to the mini-batch setting as in~\citep{nitanda:svrg,koneeny:mini}. In particular, our ASVRG method can be extended to an accelerated incremental aggregated gradient method with the SAGA estimator in \citep{defazio:saga}.

\begin{algorithm}[h]
\caption{ASVRG with $\omega=1$}
\label{alg3}
\renewcommand{\algorithmicrequire}{\textbf{Input:}}
\renewcommand{\algorithmicensure}{\textbf{Initialize:}}
\renewcommand{\algorithmicoutput}{\textbf{Output:}}
\begin{algorithmic}[1]
\REQUIRE The number of epochs $S$, the number of iterations $m$ per epoch, and the step size $\eta$.\\
\ENSURE $x^{1}_{0}=\widetilde{x}^{0}$, $m_{1}=n/4$, $\rho>1$, and the probability $P=[p_{1},\ldots,p_{n}]$.\\
\FOR{$s=1,2,\ldots,S$}
\STATE {$\widetilde{\nabla}=\frac{1}{n}\!\sum^{n}_{i=1}\!\nabla\!f_{i}(\widetilde{x}^{s-1})$;}
\FOR{$t=1,2,\ldots,m_{s}$}
\STATE {Pick $i_{t}$ from $\{1,\ldots,n\}$ randomly based on $P$;}
\STATE {$\widetilde{\nabla}\! f_{i_{t}}(x^{s}_{t-1})=\left[\nabla\! f_{i_{t}}(x^{s}_{t-1})-\nabla\! f_{i_{t}}(\widetilde{x}^{s-1})\right]/(np_{i_{t}})+\widetilde{\nabla}$;}
\STATE {$x^{s}_{t}=x^{s}_{t-1}-\eta\left[\widetilde{\nabla}\!f_{i_{t}}(x^{s}_{t-1})+\nabla\! g(x^{s}_{t-1})\right]$;}
\ENDFOR
\STATE {$\widetilde{x}^{s}=\frac{1}{m_{s}}\!\sum^{m_{s}}_{t=1}\!x^{s}_{t}$,\, $x^{s+1}_{0}=x^{s}_{m_{s}}$,\, $m_{s+1}=\min(\lfloor\rho m_{s}\rfloor,m)$;}
\ENDFOR
\OUTPUT {$\widetilde{x}^{S}$.}
\end{algorithmic}
\end{algorithm}

\subsection*{Appendix A3: Elastic-Net Regularized Logistic Regression}
In this paper, we mainly focus on the following elastic-net regularized logistic regression problem for binary classification,
\begin{equation*}
\min_{x\in\mathbb{R}^{d}}\frac{1}{n}\sum^{n}_{i=1}\log(1+\exp(-b_{i}a^{T}_{i}x))+\frac{\lambda_{1}}{2}\|x\|^{2}+\lambda_{2}\|x\|_{1},
\end{equation*}
where $\{(a_{i},b_{i})\}$ is a set of training examples, and $\lambda_{1},\lambda_{2}\!\geq\!0$ are the regularization parameters. Note that  $f_{i}(x)\!=\!\log(1\!+\!\exp(-b_{i}a^{T}_{i}x))\!+\!(\lambda_{1}/{2})\|x\|^{2}$.

In this paper, we used the two publicly available data sets in the experiments: Covtype and RCV1, as listed in Table~\ref{tab1}. For fair comparison, we implemented the state-of-the-art stochastic methods such as SAGA~\citep{defazio:saga}, SVRG~\citep{johnson:svrg}, Acc-Prox-SVRG~\citep{nitanda:svrg}, Catalyst~\citep{lin:vrsg}, and Katyusha~\citep{zhu:Katyusha}, and our ASVRG method in C++ with a Matlab interface, and conducted all the experiments on a PC with an Intel i5-4570 CPU and 16GB RAM.

\begin{table}[th]
\centering
\caption{Summary of data sets used for our experiments.}
\label{tab1}
\setlength{\tabcolsep}{15.9pt}
\renewcommand\arraystretch{1.39}
\begin{tabular}{lcc}
\hline
\ Data sets    & Covtype  & RCV1\\
\hline
\ Number of training samples, $n$ & 581,012 & 20,242 \\
\ Number of dimensions, $d$       & 54  & 47,236 \\
\ Sparsity                        & 22.12\%  & 0.16\% \\
\ Size                            & 50M & 13M\\
\hline
\end{tabular}
\end{table}


\section{Proof of Lemma 2}
Before proving the key Lemma 2, we first give the following lemma and properties, which are useful for the convergence analysis of our ASVRG method.

\begin{lemma}
\label{lemm11}
Suppose Assumption 1 holds. Then the following inequality holds
\begin{equation}
\begin{split}
&\mathbb{E}\!\left[\left\|\widetilde{\nabla}\! f_{i_{t}}\!(x^{s}_{t-1})-\nabla\! f(x^{s}_{t-1})\right\|^{2}\right]
\leq\,2\widetilde{L}\!\left(f(\widetilde{x}^{s-1})-f(x^{s}_{t-1})+\left\langle \nabla\! f(x^{s}_{t-1}),\; x^{s}_{t-1}-\widetilde{x}^{s-1}\right\rangle\right)\!,
\end{split}
\end{equation}
where $\langle\cdot,\cdot\rangle$ denotes the inner product (i.e., $\langle x,y\rangle\!=\!x^{T}y$ for all $x,y\!\in\!\mathbb{R}^{d}$), and $\widetilde{L}\!=\!\max_{j}L_{j}/(p_{j}n)$. When $p_{i}\!=\!1/n$ (i.e., uniform random sampling), $\widetilde{L}\!=\!L_{\max}\!:=\!\max_{j}L_{j}$, while $\widetilde{L}\!=\!L_{\textup{avg}}\!:=\!\frac{1}{n}\!\sum^{n}_{j=1}\!L_{j}$ when $p_{i}\!=\!L_{i}/\sum^{n}_{j=1}\!L_{j}$ (i.e., the sampling probabilities $p_{i}$ for $i\!\in\!\{1,\ldots,n\}$ are proportional to their Lipschitz constants $L_{i}$ of ${\nabla}\!f_{i}(\cdot)$).
\end{lemma}

The proof of Lemma~\ref{lemm11} is similar to that of Lemma 3.4 in \citep{zhu:Katyusha}. For the sake of completeness, we give the detailed proof of Lemma~\ref{lemm11} as follows. Their main difference is that Lemma~\ref{lemm11} provides the upper bound on the expected variance of the modified stochastic gradient estimator, i.e.,
\begin{equation*}
\widetilde{\nabla}\! f_{i_{t}}\!(x^{s}_{t-1})=\left[\nabla\! f_{i_{t}}\!(x^{s}_{t-1})-\nabla\! f_{i_{t}}\!(\widetilde{x}^{s-1})\right]\!/(np_{i_{t}})+\widetilde{\nabla},
\end{equation*}
while the upper bound in Lemma 3.4 in \citep{zhu:Katyusha} is for the standard stochastic gradient estimator in \citep{johnson:svrg,zhang:svrg}. Obviously, the upper bound in Lemma~\ref{lemm11} is much tighter than that in \citep{johnson:svrg,xiao:prox-svrg,zhu:univr}, e.g., Corollary 3.5 in \citep{xiao:prox-svrg} and Lemma A.2 in \citep{zhu:univr}.

\begin{proof}
Now we take expectations with respect to the random choice of $i_{t}$, to obtain
\begin{equation}\label{equ1110}
\begin{split}
&\;\mathbb{E}\!\left[\frac{1}{np_{i_{t}}}\!\left[\nabla\! f_{i_{t}}(x^{s}_{t-1})-\nabla\! f_{i_{t}}(\widetilde{x}^{s-1})\right]\right]\\
=&\;\sum^{n}_{i=1}\frac{p_{i}}{np_{i}}\!\left[\nabla\! f_{i}(x^{s}_{t-1})-\nabla\! f_{i}(\widetilde{x}^{s-1})\right]\\
=&\;\sum^{n}_{i=1}\frac{1}{n}\left[\nabla\! f_{i}(x^{s}_{t-1})-\nabla\! f_{i}(\widetilde{x}^{s-1})\right]\\
=&\;\nabla\! f(x^{s}_{t-1})-\nabla\! f(\widetilde{x}^{s-1}).
\end{split}
\end{equation}

Theorem 2.1.5 in \citep{nesterov:co} immediately implies the following result.
\begin{equation*}
\begin{split}
&\left\|\nabla\!f_{i}(x^{s}_{t-1})-\nabla\!f_{i}(\widetilde{x}^{s-1})\right\|^{2}
\leq 2L_{i}\!\left[f_{i}(\widetilde{x}^{s-1})-f_{i}(x^{s}_{t-1})+\langle \nabla\!f_{i}(x^{s}_{t-1}),\; x^{s}_{t-1}-\widetilde{x}^{s-1}\rangle\right]\!.
\end{split}
\end{equation*}

Dividing both sides of the above inequality by $1/(n^{2}p_{i})$, and summing it over $i\!=\!1,\ldots,n$, we obtain
\begin{equation}\label{equ1111}
\begin{split}
&\;\frac{1}{n}\sum^{n}_{i=1}\frac{1}{n p_{i}}\left\|\nabla\! f_{i}(x^{s}_{t-1})-\nabla\! f_{i}(\widetilde{x}^{s-1})\right\|^{2}\\
\leq & \;2\widetilde{L}\!\left[f(\widetilde{x}^{s-\!1})-f(x^{s}_{t-1})+\langle \nabla\! f(x^{s}_{t-1}),\; x^{s}_{t-1}-\widetilde{x}^{s-1}\rangle\right]\!.
\end{split}
\end{equation}

Using the definition of $\widetilde{\nabla}\! f_{i_{t}}(x^{s}_{t-1})=[\nabla\! f_{i_{t}}(x^{s}_{t-1})-\nabla\! f_{i_{t}}(\widetilde{x}^{s-1})]/(n p_{i_{t}})+\nabla\! f(\widetilde{x}^{s-1})$, (\ref{equ1110}), and (\ref{equ1111}), we have
\begin{equation*}
\begin{split}
&\;\mathbb{E}\!\left[\left\|\widetilde{\nabla}\! f_{i_{t}}\!(x^{s}_{t-1})-\nabla\! f(x^{s}_{t-1})\right\|^{2}\right]\\
=&\;\mathbb{E}\!\left[\left\|\nabla\! f(\widetilde{x}^{s-1})-\nabla\! f(x^{s}_{t-1})-\frac{\nabla\! f_{i_{t}}\!(\widetilde{x}^{s-1})-\nabla\! f_{i_{t}}\!(x^{s}_{t-1})}{n p_{i_{t}}}\right\|^{2}\right]\\
\leq&\; \mathbb{E}\!\left[\frac{1}{n^{2} p^{2}_{i_{t}}}\!\left\|\nabla\! f_{i_{t}}\!(x^{s}_{t-1})-\nabla\! f_{i_{t}}\!(\widetilde{x}^{s-1})\right\|^{2}\right]\\
=&\;\frac{1}{n}\sum^{n}_{i=1}\frac{1}{n p_{i}}\left\|\nabla\! f_{i}(x^{s}_{t-1})-\nabla\! f_{i}(\widetilde{x}^{s-1})\right\|^{2}\\
\leq &\; 2\widetilde{L}\left[f(\widetilde{x}^{s-1})-f(x^{s}_{t-1})+\langle \nabla\! f(x^{s}_{t-1}),\; x^{s}_{t-1}-\widetilde{x}^{s-1}\rangle\right]\!,
\end{split}
\end{equation*}
where the first inequality follows from the fact that $\mathbb{E}[\|\mathbb{E}[x]-x\|^{2}]=\mathbb{E}[\|x\|^{2}]-\|\mathbb{E}[x]\|^{2}\leq \mathbb{E}[\|x\|^{2}]$, and the second inequality holds due to (\ref{equ1111}).
\end{proof}

\begin{property}[\cite{lan:sgd}]
\label{prop2}
Assume that $z^{*}$ is an optimal solution of the following problem,
\begin{displaymath}
\min_{z}\frac{\nu}{2}\|z-z_{0}\|^{2}+h(z),
\end{displaymath}
where $h(z)$ is a convex function (but possibly non-differentiable). Then for any $z\!\in\!\mathbb{R}^{d}$,
\begin{displaymath}
h(z^{*})+\frac{\nu}{2}\|z^{*}-z_{0}\|^{2}+\frac{\nu}{2}\|z-z^{*}\|^{2}\leq h(z)+\frac{\nu}{2}\|z-z_{0}\|^{2}.
\end{displaymath}
\end{property}

\begin{property}\label{prop3}
Assume that the stochastic momentum weight $\omega_{s}$ in Algorithm 2 satisfies the following conditions:
\begin{equation}\label{equ95}
\omega_{0}\leq 1-\frac{1}{\alpha-1}\;\;\textup{and}\;\;\frac{1-\omega_{s}}{\omega^{2}_{s}}=\frac{1}{\omega^{2}_{s-1}},
\end{equation}
where $\alpha=1/(\widetilde{L}\eta)$. Then the following properties hold:
\begin{displaymath}
\begin{split}
\omega_{s}=\frac{\sqrt{\omega^{4}_{s-1}+4\omega^{2}_{s-1}}-\omega^{2}_{s-1}}{2},\quad\omega_{s}\leq\frac{2}{s+2}.
\end{split}
\end{displaymath}
\end{property}

\begin{proof}
Using the equality in \eqref{equ95}, it is easy to show that
\begin{displaymath}
\omega_{s}=\frac{\sqrt{\omega^{4}_{s-1}+4\omega^{2}_{s-1}}-\omega^{2}_{s-1}}{2}\geq0.
\end{displaymath}
In the following, we will prove by induction that $\omega_{s}\leq \frac{2}{s+2}$. Firstly, we have
\begin{displaymath}
\omega_{0}\leq 1-\frac{1}{\alpha-1}\leq 1=\frac{2}{0+2}.
\end{displaymath}
Assume that $\omega_{s-1}\leq \frac{2}{s+1}$, then we have
\begin{displaymath}
\begin{split}
\omega_{s}&=\frac{\sqrt{\omega^{4}_{s-1}+4\omega^{2}_{s-1}}-\omega^{2}_{s-1}}{2}=\frac{2}{1+\sqrt{1+\frac{4}{\omega^{2}_{s-1}}}}\\
&\leq\frac{2}{1+\sqrt{1+(s+1)^{2}}}\\
&\leq\frac{2}{s+2}.
\end{split}
\end{displaymath}
This completes the proof.
\end{proof}

\subsection*{Proof of Lemma 2:}
\begin{proof}
Let $\widetilde{\nabla}_{t}:=\!\left[\nabla f_{i_{t}}\!(x^{s}_{t-1})-\nabla f_{i_{t}}\!(\widetilde{x}^{s-1})\right]/(np_{i_{t}})+\nabla f(\widetilde{x}^{s-1})$. Suppose each component function $f_{i}(\cdot)$ is $L_{i}$-smooth, which implies that the gradient of the average function $f(x)$ is convex and also Lipschitz-continuous, i.e., there exists a Lipschitz constant $L_{f}\!>0$ such that for all $x,y\!\in\!\mathbb{R}^{d}$,
\begin{displaymath}
\|\nabla f(x)-\nabla f(y)\|\leq L_{f}\|x-y\|,
\end{displaymath}
whose equivalent form is
\begin{displaymath}
f(y)\leq f(x)+\langle\nabla f(x),\;y-x\rangle+\frac{L_{f}}{2}\|y-x\|^{2}.
\end{displaymath}

Moreover, it is easy to verify that $L_{f}\leq L_{\textup{avg}}=\frac{1}{n}\sum^{n}_{j=1}L_{j}\leq \widetilde{L}$. Let $\eta=1/(\widetilde{L}\alpha)$ and $\alpha>2$ be a suitable constant, then we have
\begin{equation}\label{equ71}
\begin{split}
f(x^{s}_{t})\leq\,& f(x^{s}_{t-1})+\left\langle\nabla f(x^{s}_{t-1}),\,x^{s}_{t}-x^{s}_{t-1}\right\rangle+\frac{L_{f}}{2}\!\left\|x^{s}_{t}-x^{s}_{t-1}\right\|^{2}\\
\leq\,& f(x^{s}_{t-1})+\left\langle\nabla f(x^{s}_{t-1}),\,x^{s}_{t}-x^{s}_{t-1}\right\rangle+\frac{\widetilde{L}}{2}\!\left\|x^{s}_{t}-x^{s}_{t-1}\right\|^{2}\\
=\,& f(x^{s}_{t-1})+\left\langle\nabla f(x^{s}_{t-1}),\,x^{s}_{t}-x^{s}_{t-1}\right\rangle+\frac{\widetilde{L}\alpha}{2}\!\left\|x^{s}_{t}-x^{s}_{t-1}\right\|^{2}-\frac{\widetilde{L}(\alpha\!-\!1)}{2}\!\left\|x^{s}_{t}-x^{s}_{t-1}\right\|^{2}\\
=\,& f(x^{s}_{t-1})+\left\langle \widetilde{\nabla}_{{t}},\,x^{s}_{t}-x^{s}_{t-1}\right\rangle+\frac{\widetilde{L}\alpha}{2}\|x^{s}_{t}-x^{s}_{t-1}\|^2-\frac{\widetilde{L}(\alpha\!-\!1)}{2}\|x^{s}_{t}-x^{s}_{t-1}\|^{2}\\
&+\left\langle\nabla f(x^{s}_{t-1})-\widetilde{\nabla}_{{t}},\,x^{s}_{t}-x^{s}_{t-1}\right\rangle.
\end{split}
\end{equation}
\begin{equation}\label{equ72}
\begin{split}
&\mathbb{E}\!\left[\left\langle\nabla\! f(x^{s}_{t-1})-\widetilde{\nabla}_{{t}},\,x^{s}_{t}-x^{s}_{t-1}\right\rangle\right]\\
\leq\,& \mathbb{E}\!\left[\frac{1}{2\widetilde{L}(\alpha\!-\!1)}\left\|\nabla\!f(x^{s}_{t-1})-\widetilde{\nabla}_{{t}}\right\|^{2}+\frac{\widetilde{L}(\alpha\!-\!1)}{2}\left\|x^{s}_{t}\!-\!x^{s}_{t-1}\right\|^{2}\right]\\
\leq\,& \frac{1}{\alpha\!-\!1}\!\left[f(\widetilde{x}^{s-1})-f(x^{s}_{t-1})+\left\langle\nabla f(x^{s}_{t-1}),\;x^{s}_{t-1}-\widetilde{x}^{s-1}\right\rangle\right]+\frac{\widetilde{L}(\alpha\!-\!1)}{2}\mathbb{E}\!\left[\left\|x^{s}_{t}\!-\!x^{s}_{t-1}\right\|^{2}\right],
\end{split}
\end{equation}
where the first inequality holds due to the Young's inequality, i.e., $a^{T}b\!\leq\!{\|a\|^2}/{(2\theta)}\!+\!{\theta\|b\|^2}/{2}$ for all $\theta\!>\!0$, and the second inequality follows from Lemma~\ref{lemm11}.

Taking the expectation over the random choice of $i_{t}$, and substituting the inequality \eqref{equ72} into the inequality \eqref{equ71}, then we have
\begin{equation}\label{equ73}
\begin{split}
\mathbb{E}[F(x^{s}_{t})]&\leq f(x^{s}_{t-1})+\mathbb{E}\!\left[\left\langle\widetilde{\nabla}_{{t}}, \,x^{s}_{t}-x^{s}_{t-1}\right\rangle+\frac{\widetilde{L}\alpha}{2}\|x^{s}_{t}-x^{s}_{t-1}\|^2+g(x^{s}_{t})\right]\\
&\quad+\frac{1}{\alpha\!-\!1}\!\left[f(\widetilde{x}^{s-1})-f(x^{s}_{t-1})+\left\langle\nabla f(x^{s}_{t-1}),\;x^{s}_{t-1}-\widetilde{x}^{s-1}\right\rangle\right]\\
&\leq f(x^{s}_{t-1})+\mathbb{E}\!\left[\omega_{s-1}\left\langle \widetilde{\nabla}_{t}, \,y^{s}_{t}-y^{s}_{t-1}\right\rangle+\frac{\widetilde{L}\alpha\omega^{2}_{s-1}}{2}\|y^{s}_{t}-y^{s}_{t-1}\|^2+\omega_{s-1}g(y^{s}_{t})\right]\\
&\quad+(1-\omega_{s-1})g(\widetilde{x}^{s-1})+\frac{1}{\alpha\!-\!1}\!\left[f(\widetilde{x}^{s-1})-f(x^{s}_{t-1})+\left\langle\nabla f(x^{s}_{t-1}),\;x^{s}_{t-1}-\widetilde{x}^{s-1}\right\rangle\right]\\
&\leq \!f(x^{s}_{t-\!1})\!+\!\mathbb{E}\!\left[\omega_{s-\!1}\!\left\langle\! \widetilde{\nabla}_{\!t}, \,x^{\star}\!\!-\!y^{s}_{t-\!1}\right\rangle\!+\!\frac{\widetilde{L}\alpha\omega^{2}_{s-\!1}}{2}(\|x^{\star}\!\!-\!y^{s}_{t-\!1}\|^2\!-\!\|x^{\star}\!\!-\!y^{s}_{t}\|^2)\!+\!\omega_{s-\!1}g(x^{\star})\right]\\
&\quad+(1-\omega_{s-1})g(\widetilde{x}^{s-1})+\frac{1}{\alpha\!-\!1}\!\left[f(\widetilde{x}^{s-1})-f(x^{s}_{t-1})+\left\langle\nabla f(x^{s}_{t-1}),\;x^{s}_{t-1}-\widetilde{x}^{s-1}\right\rangle\right]\\
&= f(x^{s}_{t-\!1})+\mathbb{E}\!\left[\frac{\widetilde{L}\alpha\omega^{2}_{s-\!1}}{2}\!\left(\|x^{\star}\!-\!y^{s}_{t-1}\|^2\!-\!\|x^{\star}\!-\!y^{s}_{t}\|^2\right)\!+\!\omega_{s-1}g(x^{\star})\right]\!+\!(1\!-\!\omega_{s-\!1})g(\widetilde{x}^{s-\!1})\\
&\quad+\left\langle\nabla f(x^{s}_{t-\!1}),\;\omega_{s-1}x^{\star}+(1-\omega_{s-1})\widetilde{x}^{s-1}\!-\!x^{s}_{t-\!1}\!+\!\frac{1}{\alpha\!-\!1}(x^{s}_{t-\!1}\!-\!\widetilde{x}^{s-\!1})\right\rangle\\
&\quad+\!\mathbb{E}\!\left[\left\langle-\!\nabla\! f_{i_{t}}\!(\widetilde{x}^{s-\!1})\!+\!\nabla\! f(\widetilde{x}^{s-\!1}),\;\omega_{s-\!1}x^{\star}\!\!+\!(1\!-\!\omega_{s-\!1})\widetilde{x}^{s-1}\!\!-\!x^{s}_{t-\!1}\right\rangle\right]\!+\!\frac{f(\widetilde{x}^{s-\!1})\!-\!f(x^{s}_{t-\!1})}{\alpha\!-\!1}\\
&= f(x^{s}_{t-\!1})\!+\!\mathbb{E}\!\left[\frac{\widetilde{L}\alpha\omega^{2}_{s-\!1}}{2}\!\left(\|x^{\star}\!-\!y^{s}_{t-1}\|^2\!-\!\|x^{\star}\!-\!y^{s}_{t}\|^2\right)\!+\!\omega_{s-\!1}g(x^{\star})\right]\!+\!(1\!-\!\omega_{s-\!1})g(\widetilde{x}^{s-\!1})\\
&\quad+\left\langle\nabla f(x^{s}_{t-\!1}),\;\omega_{s-1}x^{\star}+(1-\omega_{s-1})\widetilde{x}^{s-1}\!-\!x^{s}_{t-\!1}\!+\!\frac{1}{\alpha\!-\!1}(x^{s}_{t-\!1}\!-\!\widetilde{x}^{s-\!1})\right\rangle\\
&\quad+\frac{1}{\alpha\!-\!1}\left(f(\widetilde{x}^{s-1})-f(x^{s}_{t-1})\right),
\end{split}
\end{equation}
where the first inequality holds due to the inequalities \eqref{equ71} and \eqref{equ72}; the second inequality follows from the facts that $x^{s}_{t}=\widetilde{x}^{s-1}+\omega_{s-1}(y^{s}_{t}-\widetilde{x}^{s-1})=\omega_{s-1}y^{s}_{t}+(1-\omega_{s-1})\widetilde{x}^{s-1}$, $x^{s}_{t}-x^{s}_{t-1}=\omega_{s-1}(y^{s}_{t}-y^{s}_{t-1})$, and
\begin{displaymath}
g(\omega_{s-1}y^{s}_{t}+(1-\omega_{s-1})\widetilde{x}^{s-1})\leq \omega_{s-1}g(y^{s}_{t})+(1-\omega_{s-1})g(\widetilde{x}^{s-1}).
\end{displaymath}
Since $y^{s}_{t}$ is the optimal solution of the problem (5), the third inequality follows from Property~\ref{prop2} with $z^{*}=y^{s}_{t}$, $z=x^{\star}$, $z_{0}=y^{s}_{t-1}$, $\nu=\widetilde{L}\alpha\omega_{s-1}={\omega_{s-\!1}}/{\eta}$ and $h(y):=\langle \widetilde{\nabla}_{{t}},\, y-y^{s}_{t-1}\rangle+g(y)$. The first equality holds due to the facts that
\begin{displaymath}
\begin{split}
&\;\omega_{s-1}\left\langle \widetilde{\nabla}_{{t}},\;x^{\star}-y^{s}_{t-1}\right\rangle\\
=&\;\left\langle \widetilde{\nabla}_{{t}},\;\omega_{s-1}x^{\star}+(1-\omega_{s-1})\widetilde{x}^{s-1}-x^{s}_{t-1}\right\rangle\\
=&\;\left\langle\nabla\! f_{i_{t}}\!(x^{s}_{t-\!1}),\;\omega_{s-\!1}x^{\star}\!+\!(1\!-\!\omega_{s-\!1})\widetilde{x}^{s-\!1}\!-\!x^{s}_{t-\!1}\right\rangle\\
&\;+\left\langle-\nabla\! f_{i_{t}}\!(\widetilde{x}^{s-\!1})\!+\!\nabla\! f(\widetilde{x}^{s-\!1}),\;\omega_{s-\!1}x^{\star}\!+\!(1\!-\!\omega_{s-\!1})\widetilde{x}^{s-\!1}\!-\!x^{s}_{t-\!1}\right\rangle,
\end{split}
\end{displaymath}
and $\mathbb{E}[\nabla\! f_{i_{t}}(x^{s}_{t-1})]=\nabla\! f(x^{s}_{t-1})$, and the last equality follows from the fact that
\begin{displaymath}
\mathbb{E}\left[\langle-\nabla \! f_{i_{t}}(\widetilde{x}^{s-1})+\nabla\! f(\widetilde{x}^{s-1}),\,\omega_{s-1}x^{\star}+(1-\omega_{s-1})\widetilde{x}^{s-1}-x^{s}_{t-1}\rangle\right]=0.
\end{displaymath}

Furthermore,
\begin{equation}\label{equ75}
\begin{split}
&\left\langle\nabla f(x^{s}_{t-1}),\;(1-\omega_{s-1})\widetilde{x}^{s-1}+\omega_{s-1}x^{\star}-x^{s}_{t-1}+\frac{1}{\alpha\!-\!1}(x^{s}_{t-1}-\widetilde{x}^{s-1})\right\rangle\\
=&\, \left\langle\nabla f(x^{s}_{t-1}),\;\omega_{s-1}x^{\star}+(1-\omega_{s-1}-\frac{1}{\alpha\!-\!1})\widetilde{x}^{s-1}+\frac{1}{\alpha\!-\!1}x^{s}_{t-1}-x^{s}_{t-1}\right\rangle\\
\leq\, &f\!\left(\omega_{s-1}x^{\star}+(1-\omega_{s-1}-\frac{1}{\alpha\!-\!1})\widetilde{x}^{s-1}+\frac{1}{\alpha\!-\!1}x^{s}_{t-1}\right)-f(x^{s}_{t-1})\\
\leq\, &\omega_{s-1}f(x^{\star})+\left(1-\omega_{s-1}-\frac{1}{\alpha\!-\!1}\right)f(\widetilde{x}^{s-1})+\frac{1}{\alpha\!-\!1}f(x^{s}_{t-1})-f(x^{s}_{t-1}),
\end{split}
\end{equation}
where the first inequality holds due to the fact that $\langle \nabla f(x),\,y-x\rangle\leq f(y)-f(x)$, and the last inequality follows from the convexity of the function $f(\cdot)$ and the assumption that $1-\omega_{s-1}-\frac{1}{\alpha-1}=1-\omega_{s-1}-\frac{\widetilde{L}\eta}{1-\widetilde{L}\eta}\geq0$. Substituting the inequality \eqref{equ75} into the inequality \eqref{equ73}, we have
\begin{equation*}\label{equ76}
\begin{split}
\mathbb{E}\!\left[F(x^{s}_{t})\right]\leq&\, f(x^{s}_{t-1})+\mathbb{E}\!\left[\frac{\widetilde{L}\alpha\omega^{2}_{s-\!1}}{2}\!\!\left(\|x^{\star}\!-\!y^{s}_{t-\!1}\|^2\!-\!\|x^{\star}\!-\!y^{s}_{t}\|^2\right)\!+\!\omega_{s-\!1}g(x^{\star})\!+\!(1\!-\!\omega_{s-1})g(\widetilde{x}^{s-\!1})\right]\\
&\,+\omega_{s-1}f(x^{\star})+\left(1-\omega_{s-1}-\frac{1}{\alpha\!-\!1}\right)f(\widetilde{x}^{s-1})+\frac{1}{\alpha\!-\!1}f(x^{s}_{t-1})-f(x^{s}_{t-1})\\
&\,+\frac{1}{\alpha\!-\!1}\left(f(\widetilde{x}^{s-1})-f(x^{s}_{t-1})\right)\\
=&\;\omega_{s-1}F(x^{\star})+(1-\omega_{s-1})F(\widetilde{x}^{s-1})+\frac{\widetilde{L}\alpha\omega^{2}_{s-1}}{2}\mathbb{E}\!\left[\|x^{\star}-y^{s}_{t-1}\|^2-\|x^{\star}-y^{s}_{t}\|^2\right].
\end{split}
\end{equation*}
Therefore, we have
\begin{equation*}
\begin{split}
&\,\mathbb{E}\!\left[F(x^{s}_{t})-F(x^{\star})\right]\\
\leq&\,(1\!-\!\omega_{s-1})\mathbb{E}\!\left[[F(\widetilde{x}^{s-1})\!-F(x^{\star})\right]+\frac{\widetilde{L}\alpha\omega^{2}_{s-1}}{2}\mathbb{E}\!\left[\|x^{\star}-y^{s}_{t-1}\|^2-\|x^{\star}-y^{s}_{t}\|^2\right]\!.
\end{split}
\end{equation*}

Since
\begin{equation*}
\widetilde{x}^{s}=\frac{1}{m}\sum^{m}_{t=1}x^{s}_{t} \quad\textup{and}\quad F\!\left(\frac{1}{m}\sum^{m}_{t=1}x^{s}_{t}\right)\leq \frac{1}{m}\sum^{m}_{t=1}F(x^{s}_{t}),
\end{equation*}
by taking the expectation over the random choice of the history of random variables $i_{1},\cdots,i_{m}$ on the above inequality, and summing it over $t=1,\cdots,m$ at the $s$-th stage, then we have
\begin{equation*}
\begin{split}
&\;\mathbb{E}\!\left[F(\widetilde{x}^{s})-F(x^{\star})\right]\\
\leq&\;(1\!-\!\omega_{s-1})\mathbb{E}\!\left[F(\widetilde{x}^{s-1})-F(x^{\star})\right]+\frac{\widetilde{L}\alpha\omega^{2}_{s-1}}{2m}\mathbb{E}\!\left[\left\|x^{\star}\!-\!y^{s}_{0}\right\|^2-\left\|x^{\star}\!-\!y^{s}_{m}\right\|^2\right]\\
=&\;(1\!-\!\omega_{s-1})\mathbb{E}\!\left[F(\widetilde{x}^{s-1})-F(x^{\star})\right]+\frac{\omega^{2}_{s-1}}{2m\eta}\mathbb{E}\!\left[\left\|x^{\star}\!-\!y^{s}_{0}\right\|^2-\left\|x^{\star}\!-\!y^{s}_{m}\right\|^2\right].
\end{split}
\end{equation*}
This completes the proof.
\end{proof}

\section{}
\subsection*{Appendix C1: Proof of Theorem 3}
\begin{proof}
Since the regularizer $g(x)$ is $\mu$-strongly convex, then the objective function $F(x)$ is also strongly convex with the parameter $\widetilde{\mu}\geq \mu$, i.e.\ there exists a constant $\widetilde{\mu}>0$ such that for all $x\in \mathbb{R}^{d}$
\begin{equation*}
F(x)\geq F(x^{\star})+\xi^{T}(x-x^{\star})+\frac{\widetilde{\mu}}{2}\|x-x^{\star}\|^{2},\quad\forall\xi\in \partial F(x^{\star}),
\end{equation*}
where $\partial F(x)$ is the subdifferential of $F(\cdot)$ at $x$.

Since $0\in \partial F(x^{\star})$, then we have
\begin{equation}\label{equ77}
F(x)-F(x^{\star})\geq \frac{\widetilde{\mu}}{2}\|x-x^{\star}\|^{2}\geq \frac{\mu}{2}\|x-x^{\star}\|^{2}.
\end{equation}

Using the above inequality, Lemma 2 with $\omega_{s}=\omega$ for all stages, and $y^{s}_{0}=\widetilde{x}^{s-1}$, we have
\begin{equation*}
\begin{split}
&\mathbb{E}\!\left[F(\widetilde{x}^{s})-F(x^{\star})\right]\\
\leq&\,(1-\omega)\mathbb{E}\!\left[F(\widetilde{x}^{s-1})-F(x^{\star})\right]+\frac{\widetilde{L}\alpha\omega^{2}}{2m}\mathbb{E}\!\left[\left\|x^{\star}-y^{s}_{0}\right\|^2-\left\|x^{\star}-y^{s}_{m}\right\|^2\right]\\
\leq&\,(1-\omega)\mathbb{E}\!\left[F(\widetilde{x}^{s-1})-F(x^{\star})\right]+\frac{\widetilde{L}\alpha\omega^{2}}{\mu m}\left[F(\widetilde{x}^{s-1})-F(x^{\star})\right]\\
=&\,\left(1-\omega+\frac{\widetilde{L}\alpha\omega^{2}}{\mu m}\right)\mathbb{E}\!\left[F(\widetilde{x}^{s-1})-F(x^{\star})\right]\\
=&\,\left(1-\omega+\frac{\omega^{2}}{\mu m\eta}\right)\mathbb{E}\!\left[F(\widetilde{x}^{s-1})-F(x^{\star})\right],
\end{split}
\end{equation*}
where the first inequality holds due to Lemma 2, and the second inequality follows from the inequality in~\eqref{equ77}.

This completes the proof.
\end{proof}

\subsection*{Appendix C2: Proof of Corollary 4}

For Algorithm 1 with Option I, the theoretical suggestion of the parameter settings for the learning rate $\eta$, the momentum parameter $\omega$ and the epoch size $m$ is shown in Table~\ref{tab2}.

\begin{table}                                                                                                                                                                                                                                                             \centering                                                                                                                                                                                                                                                          \small                                                                                                                                                                                                                                                \caption{Theoretical suggestion for the parameters $\eta$, $\omega$, and $m$.}                                                                                                                                                                 \vskip 0.06in                                                                                                                                                                                                                            \label{tab2}
\renewcommand\arraystretch{1.69}                                                                                                                                                                                                            \setlength{\tabcolsep}{12pt}
\begin{tabular}{cccc}                                                                                                                                                                              \hline
Condition & Learning rate $\eta$ & Parameter $\omega$ & Epoch Length $m$\\                                                                                                                                                  \hline                                                                                                                                                                                 ${m\mu}/{\widetilde{L}} \in [0.68623, 145.72]$ & $\frac{2}{5}\sqrt{{1}/{(\mu m \widetilde{L})}}$ & $ \frac{2}{25}\sqrt{{m\mu}/{\widetilde{L}}}$ & $\Theta(n)$ \\                                                                                                                                                                                  otherwise & ${1}/{(5\widetilde{L})}$ & ${1}/{5}$ & ${2\widetilde{L}}/{\mu}$\\                                                                                                                                   \hline                                                                                                                                                                                                                  \end{tabular}
\end{table}

\vspace{2mm}
\begin{proof}
Using the inequality in Theorem 3, we have
\[
\mathbb{E}\!\left[F(\widetilde{x}^{S})-F(x^{\star})\right]\leq\left(1-\omega+\frac{\omega^{2}}{\mu m\eta}\right)^{S}[F(\widetilde{x}^{0})-F(x^{\star})].
\]
Then by setting $\eta = \sqrt{\frac{1}{a^2\mu m \widetilde{L}}}$, $\omega = \sqrt{\frac{m\mu}{b^2\widetilde{L}}}$ for some constants $a$ and $b$, $m = \Theta(n)$, we have
\[
\left(1 - \omega +  \frac{\omega^2}{\mu m\eta}\right)^S =\left(1-\frac{b-a}{b^2}\sqrt{\frac{m\mu}{\widetilde{L}}}\right)^S,
\]
which means that our algorithm needs
\[
S = O\left(\frac{b^2}{b-a}\sqrt{\frac{\widetilde{L}}{\mu n}}\right) \log{\frac{F(\tilde{x}^{0})-F(x^{\star})}{\varepsilon}},	
\]
epochs to an $\varepsilon$-suboptimal solution. Then the oracle complexity of Algorithm 1 with Option I is
\begin{equation*}
\mathcal{O}(S(m + n)) = \mathcal{O}\left(\frac{b^2}{b-a}\sqrt{\frac{n\widetilde{L} }{\mu}}\log{\frac{F(\tilde{x}^{0}) - F(x^{\star})}{\varepsilon}}\right).
\end{equation*}
	
Next we need to find the constants $a$, $b$ as well as a region for ${m\mu}/{\widetilde{L}}$ that makes the above bound valid subject to some constrains,
\begin{equation}\label{constraint}
0<\omega\leq 1-\frac{\widetilde{L}\eta}{1-\widetilde{L}\eta}.
\end{equation}
By substituting our parameter settings, we get
\[
\frac{1}{b}\sqrt{\frac{m\mu}{\widetilde{L}}} - \left(\frac{1}{ab} + 1\right) + \frac{2}{a}\sqrt{\frac{\widetilde{L}}{\mu m}}\leq 0.
\]
In order for the above inequality to has a solution, the constants $a$ and $b$ should satisfy the following inequalities:
\[
\begin{cases}
b > a > 0, \\
ab \leq 3-2\sqrt{2}\text{, \,or\, } ab \geq 3+2\sqrt{2}.
\end{cases}
\]

Suppose that the above inequalities are satisfied. Let $\zeta_1$, $\zeta_2$ with $\zeta_1 \leq \zeta_2$ be the solutions to ${x^2}/{b} - ({1}/{ab} + 1)x +{2}/{a} = 0$, if ${m\mu}/{\widetilde{L}}$ satisfies
\begin{equation}\label{kappa_n_condition}
\zeta_1^2 \leq \frac{m\mu}{\widetilde{L}} \leq \zeta_2^2,
\end{equation}
then the oracle complexity in this case is $\mathcal{O}\left(\sqrt{{n\widetilde{L}}/{\mu}}\log{\frac{F(\tilde{x}^{0}) - F(x^{\star})}{\varepsilon}}\right)$.
	
For example, let $a = 2.5$, $b = 12.5$, then the range in~\eqref{kappa_n_condition} is from approximately $0.68623$ to $145.72$, that is, ${m\mu}/{\widetilde{L}} \in [0.68623, 145.72]$.
	
Now we consider the other case, i.e., out of the range in~\eqref{kappa_n_condition}. Setting $\omega = {1}/{5}$, $\eta = {1}/{(5\widetilde{L})}$, $m={2\widetilde{L}}/{\mu}$ (one can easily verify that this setting satisfies the constraint in~\eqref{constraint}), we have
\[
1 - \omega +  \frac{\omega^2}{\mu m\eta} = 0.9.
\]
Thus, the oracle complexity for this case is $\mathcal{O}\left((n+{\widetilde{L}}/{\mu})\log{\frac{F(\tilde{x}^{0}) - F(x^{\star})}{\varepsilon}}\right)$.
\end{proof}

\subsection*{Appendix C3: Proof of Corollary 5}

For Algorithm 1 with Option II, the theoretical suggestion of the parameter settings for the learning rate $\eta$, the momentum parameter $\omega$ and the epoch size $m$ is shown in Table~\ref{tab3}.

\begin{table}
\centering
\small
\caption{Theoretical suggestion for the parameters $\eta$, $\omega$, and $m$.}
\vskip 0.06in                                                                                                                                                                                                                            \label{tab3}
\renewcommand\arraystretch{1.69}
\setlength{\tabcolsep}{16pt}
\begin{tabular}{cccc}
\hline
Condition &Learning rate $\eta$ & Parameter $\omega$ & Epoch Length $m$\\
\hline
${m\mu}/{\widetilde{L}} \leq {3}/{4}$ & ${1}/(3\widetilde{L})$  & $\sqrt{{(m\mu)}/{(3\widetilde{L}})}$  &$\Theta(n)$ \\
${m\mu}/{\widetilde{L}} > {3}/{4}$  & ${1}/{(4m\mu)}$  &${1}/{2}$   &$\Theta(n)$\\
\hline
\end{tabular}
\end{table}

\vspace{2mm}
\begin{proof}
Using Lemma 2 and $\omega_{s}\equiv\omega$, we have
\[
\mathbb{E}\!\left[F(\widetilde{x}^{s})-F(x^{\star})\right]
\leq\,(1-\omega)\mathbb{E}\!\left[F(\widetilde{x}^{s-1})-F(x^{\star})\right]+\frac{\omega^{2}}{2\eta m}\mathbb{E}\!\left[\left\|x^{\star}\!-y^{s}_{0}\right\|^2-\left\|x^{\star}\!-y^{s}_{m}\right\|^2\right]\!.
\]
	
Let $\tilde{\Delta}_s= F(\tilde{x}^s) - F(x^{\star})$, $\Lambda^{s}_{t} = \|x^{\star}-y^s_{t}\|^2$, the above inequality becomes
\begin{equation*}
\mathbb{E}\!\left[\tilde{\Delta}_s\right] \leq (1-\omega)\mathbb{E}\!\left[\tilde{\Delta}_{s-1}\right] + \frac{\omega^{2}}{2\eta m}\mathbb{E}\left[\Lambda^s_0  - \Lambda^s_m\right].
\end{equation*}
Subtracting $(1\!-\!\omega)\mathbb{E}[\tilde{\Delta}_{s}]$ to both sides of the above inequality, we can rewrite the inequality as
\[
\begin{aligned}
\mathbb{E}\left[\tilde{\Delta}_s\right] \leq \frac{1 - \omega}{\omega}\mathbb{E}\left[\tilde{\Delta}_{s-1} - \tilde{\Delta}_s\right] + \frac{\omega}{2\eta m}\mathbb{E}\left[\Lambda^s_0  - \Lambda^s_m\right].
\end{aligned}
\]
Assume that our algorithm needs to restart every $\mathcal{S}$ epochs. Then in $\mathcal{S}$ epochs, by summing the above inequality overt $s = 1\ldots \mathcal{S}$, we have
\[
\begin{aligned}
\sum_{s=1}^{\mathcal{S}}\mathbb{E}\left[\tilde{\Delta}_s\right] &\leq \frac{1 - \omega}{\omega}\mathbb{E}\left[\tilde{\Delta}_{0} - \tilde{\Delta}_{\mathcal{S}}\right] + \frac{\omega}{2\eta m}\mathbb{E}\left[\Lambda^1_0  - \Lambda^\mathcal{S}_m\right] \\
&\leq \Big(\frac{1 - \omega}{\omega} + \frac{\omega}{\eta m \mu}\Big) \tilde{\Delta}_0,
\end{aligned}
\]
where the last inequality holds due to the $\mu$-strongly convex property of Problem (1). Choosing the initial vector as $x^{new}_0=\frac{1}{\mathcal{S}}\sum_{s=1}^{\mathcal{S}} {\tilde{x}^s}$ for the restart, we have
\[
\tilde{\Delta}^{new}_0 \leq \frac{\frac{1 - \omega}{\omega} + \frac{\omega}{\eta m \mu}}{\mathcal{S}} \tilde{\Delta}_0.
\]

By setting $\mathcal{S} = 2\cdot\big(\frac{1 - \omega}{\omega} + \frac{\omega}{\eta m \mu}\big)$, we have that $\tilde{\Delta}_0$ decreases by a factor of ${1}/{2}$ every $\mathcal{S}$ epochs. So in order to achieve an $\varepsilon$-suboptimal solution, the algorithm needs to perform totally $\mathcal{O}(\log{\frac{\tilde{\Delta}_0}{\varepsilon}})$ rounds of $\mathcal{S}$ epochs.
	
(I) We consider the first case, i.e., ${m\mu}/{\widetilde{L}}\!\leq {3}/{4}$. Setting $m=\Theta(n)$, $\eta={1}/{(3\widetilde{L})}$ and $\omega \!=\!\sqrt{(m\mu)/(3\widetilde{L})} \leq {1}/{2}$ (which satisfy the constraint in~\eqref{constraint}), we have $\mathcal{S} = O(\sqrt{{\widetilde{L}}/({n\mu})})$, and then the oracle complexity of our algorithm is
\[
\begin{aligned}
\mathcal{O}\left(\mathcal{S}\cdot O\left(\log{\frac{F(\tilde{x}^0) - F(x^{\star})}{\varepsilon}}\right)\cdot (m + n)\right) = \mathcal{O}\left(\sqrt{\frac{n\widetilde{L}}{\mu}}\log{\frac{F(\tilde{x}^0) - F(x^{\star})}{\varepsilon}}\right).
\end{aligned}
\]

(II) We then consider the other case, i.e., ${m\mu}/{\widetilde{L}} > {3}/{4}$. Setting $m=\Theta(n)$, $\eta = {1}/{(4m\mu)} < {1}/{(3\widetilde{L})}$ and $\omega = {1}/{2}$ (which satisfy constraint in~\eqref{constraint}), we have $\mathcal{S} = 6 \in O(1)$. Therefore, the oracle complexity of our algorithm in this case is
\[
\mathcal{O}\left(n\log{\frac{F(\tilde{x}^0) - F(x^{\star})}{\varepsilon}}\right).
\]

In short, all the results imply that the oracle complexity of Algorithm 1 is
\[
\mathcal{O}\left((n\!+\!\sqrt{{n\widetilde{L}}/{\mu}})\log{\frac{F(\tilde{x}^{0}) - F(x^{\star})}{\varepsilon}}\right).
\]
This completes the proof.
\end{proof}

\section{Proof of Theorem 7}

\begin{proof}
Using Lemma 2, we have
\begin{equation*}
\begin{split}
\frac{1}{\omega^{2}_{s-1}}\mathbb{E}\!\left[F(\widetilde{x}^{s})-F(x^{\star})\right]
\leq&\,\frac{1-\omega_{s-1}}{\omega^{2}_{s-1}}\mathbb{E}\!\left[F(\widetilde{x}^{s-1})-F(x^{\star})\right]+\frac{\widetilde{L}\alpha}{2m}\mathbb{E}\!\left[\left\|x^{\star}\!-y^{s}_{0}\right\|^2-\left\|x^{\star}\!-y^{s}_{m}\right\|^2\right]\!,
\end{split}
\end{equation*}
for all $s=1,\ldots,S$. By the update rules $y^{s}_{0}=y^{s-1}_{m}$ and $(1-\omega_{s})/\omega^{2}_{s}=1/\omega^{2}_{s-1}$, and summing the above inequality over $s=1,2,\cdots,S$, we have
\begin{equation*}
\begin{split}
\frac{1}{\omega^{2}_{S-1}}\mathbb{E}\!\left[F(\widetilde{x}^{S})-F(x^{\star})\right]
\leq&\,\frac{1-\omega_{0}}{\omega^{2}_{0}}\left[F(\widetilde{x}^{0})-F(x^{\star})\right]+\frac{\widetilde{L}\alpha}{2m}\mathbb{E}\!\left[\left\|x^{\star}-y^{0}_{0}\right\|^2-\left\|x^{\star}-y^{S}_{m}\right\|^2\right]\!.
\end{split}
\end{equation*}

Using Property~\ref{prop3}, we have
\begin{equation*}
\omega_{s}\leq\frac{2}{s+2}\quad \textrm{and}\quad \omega_{0}=1-\frac{\widetilde{L}\eta}{1-\widetilde{L}\eta}=1-\frac{1}{\alpha-1},
\end{equation*}
where $\alpha={1}/{(\widetilde{L}\eta)}$. Then
\begin{equation*}
\begin{split}
&\,\mathbb{E}\!\left[F(\widetilde{x}^{S})-F(x^{\star})\right]\\
\leq&\,\frac{4(\alpha-1)}{(\alpha-2)^{2}(S+1)^{2}}\left[F(\widetilde{x}^{0})-F(x^{\star})\right]+\frac{2\widetilde{L}\alpha}{m(S+1)^{2}}\mathbb{E}\!\left[\left\|x^{\star}-y^{0}_{0}\right\|^2-\left\|x^{\star}-y^{S}_{m}\right\|^2\right]\\
\leq&\,\frac{4(\alpha-1)}{(\alpha-2)^{2}(S+1)^{2}}\left[F(\widetilde{x}^{0})-F(x^{\star})\right]+\frac{2}{m\eta(S+1)^{2}}\mathbb{E}\!\left[\left\|x^{\star}-\widetilde{x}^{0}\right\|^2\right].
\end{split}
\end{equation*}
This completes the proof.
\end{proof}
\vspace{2mm}

\section{Proof of Lemma 9}
Before proving Lemma 9, we first give the following lemma~\citep{koneeny:mini}.

\begin{lemma}
\label{lemm2}
Let $\xi_{i}\!\in\!\mathbb{R}^{d}$ for all $i\!=\!1,2,\ldots,n$, and $\bar{\xi}\!:=\!\frac{1}{n}\sum^{n}_{i=1}\xi_{i}$. $b$ is the size of the mini-batch $I_{t}$, which is chosen independently and uniformly at random from all subsets of $[n]$. Then we have
\begin{equation*}
\mathbb{E}\!\left[\left\|\frac{1}{b}\sum_{i\in{I}_{t}}\xi_{i}-\bar{\xi}\right\|^{2}\right]\leq\frac{n-b}{nb(n-1)}\sum^{n}_{i=1}\left\|\xi_{i}\right\|^{2}.
\end{equation*}
\end{lemma}
\vspace{2mm}

\subsection*{Proof of Lemma 2:}
\begin{proof}
We extend the upper bound on the expected variance of the modified stochastic gradient estimator in Lemma~\ref{lemm11} to the mini-batch setting, i.e., $b\geq2$.
\begin{displaymath}
\begin{split}
&\;\mathbb{E}\!\left[\left\|\widetilde{\nabla}\! f_{I_{t}}\!(x^{s}_{t-1})-\nabla\! f(x^{s}_{t-1})\right\|^{2}\right]\\
=&\;\mathbb{E}\!\left[\left\|\frac{1}{b}\!\sum_{i\in{I}_{t}}\!\left[\nabla\! f_{i}(x^{s}_{t-1})-\nabla\! f_{i}(\widetilde{x}^{s-1})\right]/(np_{i})+\nabla\! f(\widetilde{x}^{s-1})-\nabla\! f(x^{s}_{t-1})\right\|^{2}\right]\\
\leq&\;\frac{n\!-\!b}{b(n\!-\!1)}\frac{1}{n}\sum^{n}_{i=1}\frac{1}{np_{i}}\left\|\nabla\! f_{i}(x^{s}_{t-1})-\nabla\! f_{i}(\widetilde{x}^{s-1})\right\|^{2}\\
\leq&\;\frac{2\widetilde{L}(n-b)}{b(n-1)}\left[f(\widetilde{x}^{s-1})-f(x^{s}_{t-1})-\left\langle \nabla\! f(x^{s}_{t-1}),\;\widetilde{x}^{s-1}-x^{s}_{t-1}\right\rangle\right],
\end{split}
\end{displaymath}
where the first inequality follows from Lemma~\ref{lemm2}, and the second inequality holds due to Theorem 2.1.5 in~\cite{nesterov:co}, i.e.,
\begin{displaymath}
\left\|\nabla\! f_{i}(x^{s}_{t-1})-\nabla\! f_{i}(\widetilde{x}^{s-1})\right\|^{2}\leq2L_{i}\!\left[f_{i}(\widetilde{x}^{s-1})-f_{i}(x^{s}_{t-1})-\left\langle \nabla\! f_{i}(x^{s}_{t-1}),\;\widetilde{x}^{s-1}-x^{s}_{t-1}\right\rangle\right].
\end{displaymath}
This completes the proof.
\end{proof}

\section{Proof of Theorem 10}
The proof of Theorem 10 is similar to that of Theorem 7. Hence, we briefly sketch the proof of Theorem 10 for the sake of completeness.

\begin{proof}
Let
\begin{displaymath}
\omega_{0}=1-\frac{\tau(b)\widetilde{L}\eta}{1-\widetilde{L}\eta}=1-\frac{\tau(b)}{\alpha-1},
\end{displaymath}
where $\alpha=\frac{1}{\widetilde{L}\eta}$, and $y^{0}_{0}=\widetilde{x}^{0}$, then we have
\begin{equation*}
\begin{split}
&\,\mathbb{E}\!\left[F(\widetilde{x}^{s})-F(x^{\star})\right]\\
\leq&\,\frac{4(\alpha-1)\tau(b)}{(\alpha-1-\tau(b))^{2}(s+1)^{2}}\left[F(\widetilde{x}^{0})-F(x^{\star})\right]+\frac{2\widetilde{L}\alpha}{m(s+1)^{2}}\mathbb{E}\!\left[\left\|x^{\star}-\widetilde{x}^{0}\right\|^2-\left\|x^{\star}-y^{s}_{m}\right\|^2\right]\\
\leq&\,\frac{4(\alpha-1)\tau(b)}{(\alpha-1-\tau(b))^{2}(s+1)^{2}}\left[F(\widetilde{x}^{0})-F(x^{\star})\right]+\frac{2}{\eta m(s+1)^{2}}\mathbb{E}\!\left[\left\|x^{\star}-\widetilde{x}^{0}\right\|^2\right].
\end{split}
\end{equation*}
This completes the proof.
\end{proof}

\bibliography{icml2018}

\begin{thebibliography}{55}
\providecommand{\natexlab}[1]{#1}
\providecommand{\url}[1]{\texttt{#1}}
\expandafter\ifx\csname urlstyle\endcsname\relax
  \providecommand{\doi}[1]{doi: #1}\else
  \providecommand{\doi}{doi: \begingroup \urlstyle{rm}\Url}\fi

\bibitem[Allen-Zhu(2018)]{zhu:Katyusha}
Z.~Allen-Zhu.
\newblock Katyusha: The first direct acceleration of stochastic gradient
  methods.
\newblock \emph{J. Mach. Learn. Res.}, 18:\penalty0 1--51, 2018.

\bibitem[Allen-Zhu and Hazan(2016)]{zhu:box}
Z.~Allen-Zhu and E.~Hazan.
\newblock Optimal black-box reductions between optimization objectives.
\newblock In \emph{NIPS}, pages 1606--1614, 2016.

\bibitem[Allen-Zhu and Yuan(2016)]{zhu:univr}
Z.~Allen-Zhu and Y.~Yuan.
\newblock Improved {SVRG} for non-strongly-convex or sum-of-non-convex
  objectives.
\newblock In \emph{ICML}, pages 1080--1089, 2016.

\bibitem[Babanezhad et~al.(2015)Babanezhad, Ahmed, Virani, Schmidt, Konecny,
  and Sallinen]{babanezhad:vrsg}
R.~Babanezhad, M.~O. Ahmed, A.~Virani, M.~Schmidt, J.~Konecny, and S.~Sallinen.
\newblock Stop wasting my gradients: Practical {SVRG}.
\newblock In \emph{NIPS}, pages 2242--2250, 2015.

\bibitem[Bauschke and Combettes(2011)]{bauschke:book}
H.~H. Bauschke and P.~L. Combettes.
\newblock \emph{Convex analysis and monotone operator theory in {H}ilbert
  spaces}.
\newblock CMS Books in Mathematics, Springer, 2011.

\bibitem[Beck and Teboulle(2009)]{beck:fista}
A.~Beck and M.~Teboulle.
\newblock A fast iterative shrinkage-thresholding algorithm for linear inverse
  problems.
\newblock \emph{SIAM J. Imaging Sci.}, 2\penalty0 (1):\penalty0 183--202, 2009.

\bibitem[Defazio(2016)]{defazio:sagab}
A.~Defazio.
\newblock A simple practical accelerated method for finite sums.
\newblock In \emph{NIPS}, pages 676--684, 2016.

\bibitem[Defazio et~al.(2014{\natexlab{a}})Defazio, Bach, and
  Lacoste-Julien]{defazio:saga}
A.~Defazio, F.~Bach, and S.~Lacoste-Julien.
\newblock {SAGA}: A fast incremental gradient method with support for
  non-strongly convex composite objectives.
\newblock In \emph{NIPS}, pages 1646--1654, 2014{\natexlab{a}}.

\bibitem[Defazio et~al.(2014{\natexlab{b}})Defazio, Caetano, and
  Domke]{defazio:Finito}
A.~J. Defazio, T.~S. Caetano, and J.~Domke.
\newblock Finito: A faster, permutable incremental gradient method for big data
  problems.
\newblock In \emph{ICML}, pages 1125--1133, 2014{\natexlab{b}}.

\bibitem[Flammarion and Bach(2015)]{flammarion:average}
N.~Flammarion and F.~Bach.
\newblock From averaging to acceleration, there is only a step-size.
\newblock In \emph{COLT}, pages 658--695, 2015.

\bibitem[Frostig et~al.(2015)Frostig, Ge, Kakade, and Sidford]{frostig:sgd}
R.~Frostig, R.~Ge, S.~M. Kakade, and A.~Sidford.
\newblock Un-regularizing: approximate proximal point and faster stochastic
  algorithms for empirical risk minimization.
\newblock In \emph{ICML}, pages 2540--2548, 2015.

\bibitem[Garber et~al.(2016)Garber, Hazan, Jin, Kakade, Musco, Netrapalli, and
  Sidford]{garber:svd}
D.~Garber, E.~Hazan, C.~Jin, S.~M. Kakade, C.~Musco, P.~Netrapalli, and
  A.~Sidford.
\newblock Faster eigenvector computation via shift-and-invert preconditioning.
\newblock In \emph{ICML}, pages 2626--2634, 2016.

\bibitem[Hazan and Luo(2016)]{hazan:svrf}
E.~Hazan and H.~Luo.
\newblock Variance-reduced and projection-free stochastic optimization.
\newblock In \emph{ICML}, pages 1263--1271, 2016.

\bibitem[Hien et~al.(2018)Hien, Lu, Xu, and Feng]{hien:asmd}
L.~T.~K. Hien, C.~Lu, H.~Xu, and J.~Feng.
\newblock Accelerated stochastic mirror descent algorithms for composite
  non-strongly convex optimization.
\newblock \emph{arXiv:1605.06892v5}, 2018.

\bibitem[Johnson and Zhang(2013)]{johnson:svrg}
R.~Johnson and T.~Zhang.
\newblock Accelerating stochastic gradient descent using predictive variance
  reduction.
\newblock In \emph{NIPS}, pages 315--323, 2013.

\bibitem[Kone\v{c}n\'{y} et~al.(2016)Kone\v{c}n\'{y}, Liu, Richt\'{a}rik, , and
  Tak\'{a}\v{c}]{koneeny:mini}
J.~Kone\v{c}n\'{y}, J.~Liu, P.~Richt\'{a}rik, , and M.~Tak\'{a}\v{c}.
\newblock Mini-batch semi-stochastic gradient descent in the proximal setting.
\newblock \emph{IEEE J. Sel. Top. Sign. Proces.}, 10\penalty0 (2):\penalty0
  242--255, 2016.

\bibitem[Krizhevsky et~al.(2012)Krizhevsky, Sutskever, and
  Hinton]{krizhevsky:deep}
A.~Krizhevsky, I.~Sutskever, and G.~E. Hinton.
\newblock Image{N}et classification with deep convolutional neural networks.
\newblock In \emph{NIPS}, pages 1097--1105, 2012.

\bibitem[Lan(2012)]{lan:sgd}
G.~Lan.
\newblock An optimal method for stochastic composite optimization.
\newblock \emph{Math. Program.}, 133:\penalty0 365--397, 2012.

\bibitem[Lan and Zhou(2018)]{lan:rpdg}
G.~Lan and Y.~Zhou.
\newblock An optimal randomized incremental gradient method.
\newblock \emph{Math. Program.}, 171:\penalty0 167--215, 2018.

\bibitem[Lee et~al.(2017)Lee, Lin, Ma, and Yang]{lee:dsgd}
J.~D. Lee, Q.~Lin, T.~Ma, and T.~Yang.
\newblock Distributed stochastic variance reduced gradient methods by sampling
  extra data with replacement.
\newblock \emph{J. Mach. Learn. Res.}, 18:\penalty0 1--43, 2017.

\bibitem[Lin et~al.(2015)Lin, Mairal, and Harchaoui]{lin:vrsg}
H.~Lin, J.~Mairal, and Z.~Harchaoui.
\newblock A universal catalyst for first-order optimization.
\newblock In \emph{NIPS}, pages 3366--3374, 2015.

\bibitem[Liu et~al.(2017)Liu, Shang, and Cheng]{liu:sadmm}
Y.~Liu, F.~Shang, and J.~Cheng.
\newblock Accelerated variance reduced stochastic {ADMM}.
\newblock In \emph{AAAI}, pages 2287--2293, 2017.

\bibitem[Mahdavi et~al.(2013)Mahdavi, Zhang, and Jin]{mahdavi:sgd}
M.~Mahdavi, L.~Zhang, and R.~Jin.
\newblock Mixed optimization for smooth functions.
\newblock In \emph{NIPS}, pages 674--682, 2013.

\bibitem[Mairal(2015)]{mairal:miso}
J.~Mairal.
\newblock Incremental majorization-minimization optimization with application
  to large-scale machine learning.
\newblock \emph{SIAM J. Optim.}, 25\penalty0 (2):\penalty0 829--855, 2015.

\bibitem[Mania et~al.(2017)Mania, Pan, Papailiopoulos, Recht, Ramchandran, and
  Jordan]{mania:svrg}
H.~Mania, X.~Pan, D.~Papailiopoulos, B.~Recht, K.~Ramchandran, and M.~I.
  Jordan.
\newblock Perturbed iterate analysis for asynchronous stochastic optimization.
\newblock \emph{SIAM J. Optim.}, 27\penalty0 (4):\penalty0 2202--2229, 2017.

\bibitem[Murata and Suzuki(2017)]{murata:dual}
T.~Murata and T.~Suzuki.
\newblock Doubly accelerated stochastic variance reduced dual averaging method
  for regularized empirical risk minimization.
\newblock In \emph{NIPS}, pages 608--617, 2017.

\bibitem[Needell et~al.(2016)Needell, Srebro, and Ward]{needell:sgd}
D.~Needell, N.~Srebro, and R.~Ward.
\newblock Stochastic gradient descent, weighted sampling, and the randomized
  {K}aczmarz algorithm.
\newblock \emph{Math. Program.}, 155:\penalty0 549--573, 2016.

\bibitem[Nesterov(1983)]{nesterov:fast}
Y.~Nesterov.
\newblock A method of solving a convex programming problem with convergence
  rate ${O}(1/k^2)$.
\newblock \emph{Soviet Math. Doklady}, 27:\penalty0 372--376, 1983.

\bibitem[Nesterov(2004)]{nesterov:co}
Y.~Nesterov.
\newblock \emph{Introductory Lectures on Convex Optimization: A Basic Course}.
\newblock Kluwer Academic Publ., Boston, 2004.

\bibitem[Nesterov(2005)]{nesterov:smooth}
Y.~Nesterov.
\newblock Smooth minimization of non-smooth functions.
\newblock \emph{Math. Program.}, 103:\penalty0 127--152, 2005.

\bibitem[Nitanda(2014)]{nitanda:svrg}
A.~Nitanda.
\newblock Stochastic proximal gradient descent with acceleration techniques.
\newblock In \emph{NIPS}, pages 1574--1582, 2014.

\bibitem[Orabona et~al.(2012)Orabona, Argyriou, and Srebro]{orabona:smooth}
F.~Orabona, A.~Argyriou, and N.~Srebro.
\newblock {PRISMA}: Proximal iterative smoothing algorithm.
\newblock \emph{arXiv:1206.2372v2}, 2012.

\bibitem[Rakhlin et~al.(2012)Rakhlin, Shamir, and Sridharan]{rakhlin:sgd}
A.~Rakhlin, O.~Shamir, and K.~Sridharan.
\newblock Making gradient descent optimal for strongly convex stochastic
  optimization.
\newblock In \emph{ICML}, pages 449--456, 2012.

\bibitem[Reddi et~al.(2015)Reddi, Hefny, Sra, Poczos, and Smola]{reddi:sgd}
S.~Reddi, A.~Hefny, S.~Sra, B.~Poczos, and A.~Smola.
\newblock On variance reduction in stochastic gradient descent and its
  asynchronous variants.
\newblock In \emph{NIPS}, pages 2629--2637, 2015.

\bibitem[Roux et~al.(2012)Roux, Schmidt, and Bach]{roux:sag}
N.~Le Roux, M.~Schmidt, and F.~Bach.
\newblock A stochastic gradient method with an exponential convergence rate for
  finite training sets.
\newblock In \emph{NIPS}, pages 2672--2680, 2012.

\bibitem[Ruder(2017)]{ruder:sgd}
S.~Ruder.
\newblock An overview of gradient descent optimization algorithms.
\newblock \emph{arXiv:1609.04747v2}, 2017.

\bibitem[Schmidt et~al.(2013)Schmidt, Roux, and Bach]{schmidt:sag}
M.~Schmidt, N.~Le Roux, and F.~Bach.
\newblock Minimizing finite sums with the stochastic average gradient.
\newblock Technical report, INRIA, Paris, 2013.

\bibitem[Shalev-Shwartz and Zhang(2013)]{shalev-Shwartz:sdca}
S.~Shalev-Shwartz and T.~Zhang.
\newblock Stochastic dual coordinate ascent methods for regularized loss
  minimization.
\newblock \emph{J. Mach. Learn. Res.}, 14:\penalty0 567--599, 2013.

\bibitem[Shalev-Shwartz and Zhang(2016)]{shalev-Shwartz:acc-sdca}
S.~Shalev-Shwartz and T.~Zhang.
\newblock Accelerated proximal stochastic dual coordinate ascent for
  regularized loss minimization.
\newblock \emph{Math. Program.}, 155:\penalty0 105--145, 2016.

\bibitem[Shamir(2015)]{shamir:pca}
O.~Shamir.
\newblock A stochastic {PCA} and {SVD} algorithm with an exponential
  convergence rate.
\newblock In \emph{ICML}, pages 144--152, 2015.

\bibitem[Shamir(2016)]{shamir:wrsgd}
O.~Shamir.
\newblock Without-replacement sampling for stochastic gradient methods.
\newblock In \emph{NIPS}, pages 46--54, 2016.

\bibitem[Shamir and Zhang(2013)]{shamir:sgd}
O.~Shamir and T.~Zhang.
\newblock Stochastic gradient descent for non-smooth optimization: Convergence
  results and optimal averaging schemes.
\newblock In \emph{ICML}, pages 71--79, 2013.

\bibitem[Shang et~al.(2017)Shang, Liu, Cheng, and Zhuo]{shang:fsvrg}
F.~Shang, Y.~Liu, J.~Cheng, and J.~Zhuo.
\newblock Fast stochastic variance reduced gradient method with momentum
  acceleration for machine learning.
\newblock \emph{arXiv:1703.07948}, 2017.

\bibitem[Shang et~al.(2018{\natexlab{a}})Shang, Liu, Zhou, Cheng, Ng, and
  Yoshida]{shang:vrsgd}
F.~Shang, Y.~Liu, K.~Zhou, J.~Cheng, K.~W. Ng, and Y.~Yoshida.
\newblock Guaranteed sufficient decrease for stochastic variance reduced
  gradient optimization.
\newblock In \emph{AISTATS}, pages 1027--1036, 2018{\natexlab{a}}.

\bibitem[Shang et~al.(2018{\natexlab{b}})Shang, Zhou, Liu, Cheng, Tsang, Zhang,
  Tao, and Jiao]{shang:vrsgd1}
F.~Shang, K.~Zhou, H.~Liu, J.~Cheng, I.~W. Tsang, L.~Zhang, D.~Tao, and
  L.~Jiao.
\newblock {VR-SGD}: A simple stochastic variance reduction method for machine
  learning.
\newblock \emph{IEEE Trans. Knowl. Data Eng.}, 2018{\natexlab{b}}.

\bibitem[Sra et~al.(2016)Sra, Yu, Li, and Smola]{sra:dsgd}
S.~Sra, A.~W. Yu, M.~Li, and A.~J. Smola.
\newblock Ada{D}elay: Delay adaptive distributed stochastic optimization.
\newblock In \emph{AISTATS}, pages 957--965, 2016.

\bibitem[Su et~al.(2014)Su, Boyd, and Candes]{su:nag}
W.~Su, S.~P. Boyd, and E.~J. Candes.
\newblock A differential equation for modeling {N}esterov's accelerated
  gradient method: Theory and insights.
\newblock In \emph{NIPS}, pages 2510--2518, 2014.

\bibitem[Tseng(2010)]{tseng:sco}
P.~Tseng.
\newblock Approximation accuracy, gradient methods, and error bound for
  structured convex optimization.
\newblock \emph{Math. Program.}, 125:\penalty0 263--295, 2010.

\bibitem[Wang et~al.(2017)Wang, Wang, and Srebro]{wang:dsvrg}
J.~Wang, W.~Wang, and N.~Srebro.
\newblock Memory and communication efficient distributed stochastic
  optimization with minibatch-prox.
\newblock In \emph{COLT}, pages 1882--1919, 2017.

\bibitem[Woodworth and Srebro(2016)]{woodworth:bound}
B.~Woodworth and N.~Srebro.
\newblock Tight complexity bounds for optimizing composite objectives.
\newblock In \emph{NIPS}, pages 3639--3647, 2016.

\bibitem[Xiao and Zhang(2014)]{xiao:prox-svrg}
L.~Xiao and T.~Zhang.
\newblock A proximal stochastic gradient method with progressive variance
  reduction.
\newblock \emph{SIAM J. Optim.}, 24\penalty0 (4):\penalty0 2057--2075, 2014.

\bibitem[Zhang et~al.(2013)Zhang, Mahdavi, and Jin]{zhang:svrg}
L.~Zhang, M.~Mahdavi, and R.~Jin.
\newblock Linear convergence with condition number independent access of full
  gradients.
\newblock In \emph{NIPS}, pages 980--988, 2013.

\bibitem[Zhang(2004)]{zhang:sgd}
T.~Zhang.
\newblock Solving large scale linear prediction problems using stochastic
  gradient descent algorithms.
\newblock In \emph{ICML}, pages 919--926, 2004.

\bibitem[Zhao and Zhang(2015)]{zhao:sampling}
P.~Zhao and T.~Zhang.
\newblock Stochastic optimization with importance sampling for regularized loss
  minimization.
\newblock In \emph{ICML}, pages 1--9, 2015.

\bibitem[Zhou et~al.(2018)Zhou, Shang, and Cheng]{zhou:fsvrg}
K.~Zhou, F.~Shang, and J.~Cheng.
\newblock A simple stochastic variance reduced algorithm with fast convergence
  rates.
\newblock In \emph{ICML}, pages 5975--5984, 2018.

\end{thebibliography}

\end{document}